\documentclass[letterpaper]{article} % DO NOT CHANGE THIS
\usepackage{aaai24}  % DO NOT CHANGE THIS
\usepackage{times}  % DO NOT CHANGE THIS
\usepackage{helvet}  % DO NOT CHANGE THIS
\usepackage{courier}  % DO NOT CHANGE THIS
\usepackage[hyphens]{url}  % DO NOT CHANGE THIS
\usepackage{graphicx} % DO NOT CHANGE THIS
\def\UrlFont{\rm}  % DO NOT CHANGE THIS
\usepackage{natbib}  % DO NOT CHANGE THIS AND DO NOT ADD ANY OPTIONS TO IT
\usepackage{caption} % DO NOT CHANGE THIS AND DO NOT ADD ANY OPTIONS TO IT
\usepackage{algorithm}
\usepackage[noend]{algorithmic}
\usepackage{todonotes}

\usepackage{subcaption}

\usepackage{xcolor}
\usepackage{multirow}
\usepackage{float}
\usepackage{tabularx}
\usepackage[inline]{enumitem}
\usepackage{subdef}
\newcommand{\swap}[3][-]{312} % just an example

\usepackage{xcolor}
\usepackage{multirow}
\usepackage{float}
\usepackage{tabularx}
\usepackage{enumitem}

\newcommand{\embed}{\Phi}

\newcommand{\y}{y}

\newcommand{\indep}{\perp \!\!\! \perp}
\newcommand{\treat}{t}
\newcommand{\treatcf}{t^{\text{CF}}}

\newcommand{\adrf}{\mu}

\newcommand{\var}{\widehat{\sigma}^2}
\newcommand{\floss}{\mathcal{L}_{\text{factual}}}
\newcommand{\fctloss}{\mathcal{L}_{\text{fct}}}
\newcommand{\giloss}{\mathcal{L}_{\text{GI}}}
\newcommand{\gploss}{\mathcal{L}_{\text{KS}}}

\newcommand{\yhat}{\widehat{y}}
\newcommand{\trnD}{D}
\newcommand{\NND}{D_{\text{NN}} (\treatcf)}
\newcommand{\NNDi}{D_{\text{NN}} (\treatcf_i)}

\newcommand{\Pcf}{P^\text{CF}}
\newcommand{\gpdelta}{\epsilon_{\text{GP}}}
\usepackage{adjustbox}

\usepackage{makecell}
\usepackage{amsmath}
\usepackage{comment}

\usepackage{lipsum,booktabs}

\usepackage{xcolor,colortbl}

\definecolor{green}{rgb}{0.8,1,0.8}
\definecolor{yellow}{rgb}{1,1,0.87}
\definecolor{red}{rgb}{1,0,0}
\newcommand{\first}[1]{{\cellcolor{green} #1}}
\newcommand{\second}[1]{{\cellcolor{yellow} #1}}

\newcommand{\our}{\textsc{GIKS}}
\newcommand{\IND}{Individualized }

\usepackage{newfloat}
\usepackage{listings}
\DeclareCaptionStyle{ruled}{labelfont=normalfont,labelsep=colon,strut=off} % DO NOT CHANGE THIS
\floatstyle{ruled}
\newfloat{listing}{tb}{lst}{}
\floatname{listing}{Listing}
\title{Continuous Treatment Effect Estimation \\ Using Gradient Interpolation and Kernel Smoothing}

\author {
    Lokesh Nagalapatti,
    Akshay Iyer,
    Abir De,
    Sunita Sarawagi
}
\affiliations {
    Indian Institute of Technology Bombay\\
    nlokesh@cse.iitb.ac.in, akshaygiyer@gmail.com, abir@cse.iitb.ac.in, sunita@iitb.ac.in
}

\begin{document}

\maketitle

\begin{abstract}
  We address the \IND continuous treatment effect (ICTE) estimation problem where we predict the effect of any continuous-valued treatment on an individual using observational data.  The main challenge in this estimation task is the potential confounding of treatment assignment with an individual's covariates in the training data, whereas during inference ICTE requires prediction on independently sampled treatments. 
In contrast to prior work that relied on regularizers or unstable GAN training, we advocate the direct approach of augmenting training individuals with independently sampled treatments and inferred counterfactual outcomes.   
We infer counterfactual outcomes using a two-pronged strategy: a Gradient Interpolation for close-to-observed treatments, and a Gaussian Process based Kernel Smoothing which allows us to downweigh high variance inferences. 
  We evaluate our method on five benchmarks and show that our method outperforms six state-of-the-art methods on the counterfactual estimation error. We analyze the superior performance of our method by showing that (1) our inferred counterfactual responses are more accurate, and (2) adding them to the training data reduces the distributional distance between the confounded training distribution and test distribution where treatment is independent of covariates.  Our proposed method is model-agnostic and we show that it improves ICTE accuracy of several existing models. We release the code at: \url{https://github.com/nlokeshiisc/GIKS_release}.
\end{abstract}

\section{Introduction}
Many applications require the estimation of the effect of a continuous treatment variable on an individual's response.  For example, in healthcare we need to estimate the effect of the dose of a drug on the recovery of a patient, in economics, we need to estimate the effect of a discount on the sales of a product, and in public policy, we need to estimate the effect of income on a person's longevity.  In all these cases, observational data is available in abundance but controlled experiments to estimate these effects exactly either raise ethical issues or incur hidden costs.  The main challenge in estimating treatment effects from observational data is that in the observed data each individual is associated with one treatment dose which may be correlated with observed covariates of the individual, but during deployment, we need to estimate outcomes on {\em all} treatment doses, thereby making dose {\em independent} of the individual in the test data distribution.

Several prior work have proposed to correct the above mismatch but 
most of these have focused on binary treatments.  
Broadly, most methods rely on a combination of these three strategies:
(1) Learn shared representation of the feature with treatment specific outcome prediction~\cite{cfrnet}, (2) Impose regularizers to make feature representations distributionally independent of the treatment~\cite{cfrnet}, (3) Exploit the overlap assumption to impose instance-specific counterfactual losses in the learned feature space~\cite{bayesian_ite, overlapping_rep}.   
Recently, a subset of these strategies have been extended to the case of continuous treatments where some have focused on extending the neural architecture to handle continuous treatments \cite{DRNet,vcnet,TransTEE}, and others extend the distributional regularizers~\cite{vcnet,Bellot2022}. However, as we show in our experiment evaluation, such regularizers are not too effective in reducing \emph{individual} continuous treatment effect estimation errors.

In this paper, we propose to directly minimize counterfactual loss for each individual by inferring outcomes at independently sampled new treatments.  We infer outcomes using two types of smoothing strategies. First, by exploiting the differentiability of the 
response function to treatment, we infer the counterfactual response by gradient interpolation~\cite{gi_loss}.  Second, by exploiting the property of overlap required for the identifiability of ICTE from observational data, we infer a counterfactual response by using a Gaussian process over feature kernels.   We handle the potential unreliability of the inferred outcomes by down-weighing examples based on the variance of the GP estimate.  
%We treat these as auxiliary predictors with a few associated parameters that we fix using a validation dataset.
We show that individual-level counterfactual losses are significantly more effective in learning ICTE compared to distributional regularizers, particularly in a mini-batch setting.  We attribute the reasons for the observed gains of our method to two factors:
(1) the inferred outcomes from the data are more accurate than a baseline that is just trained on an observational dataset, and (2) the augmented data makes the training distribution closer to the test distribution.% induced during ICTE estimation where treatment is independent of the individual.

We make the following contributions in this paper:
\begin{itemize}[leftmargin=*]
     \item We address the ICTE problem by directly minimizing loss on inferred counterfactual outcomes of new treatments applied to training instances. We infer counterfactual outcomes by (1) gradient interpolation justified by the differentiability of the response function to treatments, and (2) kernel smoothing based on the overlap assumption of covariates and treatment.
     \item We evaluate our method on five benchmark data and show that we consistently outperform six existing state-of-the-art methods on ICTE.  We also demonstrate the application of our model in two medical settings.
    \item We explain the reasons for the observed gains by showing that our proximity-inferred outcomes are more accurate than the factual model, and the augmentation reduces the confounding between treatment and covariates.
    \item We show that our method is model-agnostic and provides gains on several existing model architectures.   
\end{itemize}

\section{Problem Formulation}
We use random variables: $X\in \Xcal \subset \RR^{d_x}$ for the individual's covariates, $T \in \Bcal \subset \RR$ for treatments, and $Y(\treat) \in \RR$ for potential outcomes or response when an individual is given treatment $\treat$. Our objective is to estimate the individual treatment effect $\EE[Y(\treat) | \xb]$, which represents the expected outcome when an individual $\xb \in \Xcal$ receives treatment $\treat$. In prior work, this term has also been referred to as the Average Dose-Response Function (ADRF), denoted as $\adrf(\xb, \treat)$. We adopt the Neyman–Rubin causal model~\cite{pearlbook} and estimate ADRF using the Potential Outcomes Framework. The primary challenge is to learn $\mu(\xb, \treat)$ from an observational dataset where each $\xb$ is exposed to only one treatment dose, whose selection depends on $\xb$ making the covariates correlated with the treatment.

The observational dataset $\trnD$ comprises $N$ samples $\{(\xb_i, \treat_i, y_i)\}_{i=1}^N$. Each $\xb_i \in \Xcal \subset \RR^{d_x}$ denotes the covariate observed before an individual is exposed to a treatment; $\treat_i \in \Bcal \subset \RR$ is the value of continuous treatment $T$ applied on the $i$-th instance 
and $y_i \in \RR$ captures the outcome observed for $\xb_i$ under the treatment dose $\treat_i$. We use $\treatcf_i \neq \treat_i$ to denote any new treatment that is not observed in $\trnD$.

Following prior work, to ensure the identifiability of ICTE from the observational dataset, we make the following assumptions:

\begin{enumerate}[leftmargin=0.66cm]
    \item[A1] {\em Overlap of treatment:} which states that every individual has a non-zero probability of being assigned any treatment \ie, $\Pr(\treat | \xb) \in (0, 1) \; \forall \xb \in \Xcal, \treat \in \Bcal$ 
    \item [A2] {\em identifiability of causal effect:} which states that the observed covariates $\Xcal$ block all the backdoor paths between the treatments $\Tcal$ and outcomes $Y$.
    \item[A3] {\em Differentiability of ADRF:}  We assume that ADRF $\mu(\xb, \treat)$ is differentiable w.r.t. treatment $\treat$.
\end{enumerate}

%Positivity of propensity score,} --- 
%   We make one more assumption particular to continuous treatments which is 

Assumptions A1, and A2 are standard assumptions needed for causal inference and under them, we can claim that $\EE[Y(\treat)|X]=\EE[Y|\treat,X]$.  

Following prior work ~\cite{vcnet,cfrnet,DRNet}, we model the ADRF $\mu$ using a composition of two neural networks as $\mu(\xb, \treat) = \eta(\Phi(x), t)$, where $\embed : \RR^{d_x} \rightarrow \RR^{d_e}$ embeds the covariates $\xb$, and $\eta: \RR^{d_e} \times \RR \rightarrow \RR$ predicts the average response given the embedding of an individual $\embed(\xb)$, at a treatment dose $\treat$. 
\label{sec:model_arch}  Many recent models e.g. DRNet~\cite{DRNet}, VCNet~\cite{vcnet} follow this framework.  Our default choice is to model the embedding network $\embed$ using a simple feed-forward neural network and to make the $\eta$ network sensitive to $t$ using VCNet~\cite{vcnet}, a state-of-the-art network for continuous treatment effect inference. The parameters $\theta \in \mathbb{R}^{d_\theta}$ of the $\eta$ network are obtained as predictions from another network $\aux : \mathbb{R} \rightarrow \mathbb{R}^{d_\theta}$. The $\aux$ network uses spline bases to ensure a smooth variation of $\theta$ with $t$. Thus, we can express $\eta\big(\embed(\xb), \treat\big) = \eta\big(\embed(\xb); \theta = \aux(\treat)\big)$, and the only trainable parameters of $\eta$ are the parameters of $\aux$ \ie\ the $\psi$.

The main challenge in learning the parameters of $\mu$ using observational data through standard likelihood training is the discrepancy between the training and test data distributions. The observational dataset $\trnD$ could confound treatment with $\xb$, leading to dependence between the observed treatments and $\xb$. Specifically, the training instances are drawn from $P(X, T) = P(T | X)P(X)$, where $T \not \indep X$. On the other hand, during inference, we aim to estimate ADRF for an individual under arbitrarily assigned treatments, implying that $T \indep X$. This corresponds to the test instances being drawn from $\Pcf(X, T) = P(X)P(T)$. In the following, we show how we address this issue of disparity.

 \section{\our: Our Proposed Approach}
Our main idea is to bridge the gap between the train and test distributions with inferred counterfactuals from auxiliary layers that harness data proximity. We start with a base model with $\Phi$ and $\eta$ trained  with factual loss on the training data: 
$    \min_{\Phi,\eta}\sum_{i=1}^N \ell\big( \eta(\embed(\xb_i), \treat_i), y_i \big) $.
 Note that our losses are model agnostic, they can be integrated on top of several base architectures (\cf\ Section \ref{sec:different_base}).
Thereafter,
for each instance $(\xb_i, \treat_i, y_i)$ in the training set, %we not only consider the factual loss with the given treatment $\treat_i$ and outcome $y_i$, but also 
we sample new treatments $\treatcf_i$ from $P(T)$ and 
since there is no supervision for $y_i(\treatcf_i)$, we infer pseudo-targets $\yhat_i (\treatcf_i)$ by leveraging the proximity in the $\embed(X)$ and $T$ space to other training examples in two ways: First, we use the ADRF differentiability assumption (A4) to predict responses for new treatments that are within a small distance $\delta$ of the observed treatment using Taylor's expansion of $\eta$, and impose a loss, which we call the GI loss $\giloss$. Second, for treatments with a larger distance $|\treatcf_i - \treat_i| > \delta$, we rely on the overlap assumption (A1) and employ kernel smoothing in the embedding space $\embed(\xb)$ over samples $\xb_j$ in $D$, whose observed treatments are close to $\treatcf_i$.
This gives us an inferred $\yhat_i(\treatcf_i)$ and variance $\var(\yhat(\treatcf_i))$. We use these to impose a confidence-weighted loss which we call $\gploss$.  % to impose a loss on $\eta(\embed(\xb_i), \treatcf_i)$.  
We elaborate on these two losses:

%%TODOSS: 
\subsection{Gradient Interpolated Inferred Counterfactual Outcomes}
Since, we assume that the learned ADRF $\mu(\xb_i, \treat_i) = \eta(\embed(\xb_i), \treat_i)$ is differentiable w.r.t. $\treat$, for any new $\treatcf_i$ that lies close to $\treat_i$, we can infer its response using  a first-order Taylor series expansion of $\mu$ around $\treatcf_i$, and use these to impose instance-specific counterfactual loss as follows:
\begin{align}
    \min_{\embed, \eta} \giloss = \sum_{i=1}^N \mathcal{L}(\eta(\embed(\xb_i), \treatcf_i), \yhat_i(\treatcf_i)) \\
~~~\text{where}~~  \yhat_i(\treatcf_i) =
                   y_i - (\treat_i - \treatcf_i ) \frac{\partial \eta(\embed(\xb_i),\treat)}{\partial \treat} \label{eq:giloss}
\end{align}

Despite the spline parameterization of the VCNet, we show in Figure \ref{fig:ihdp_smooth}, and Table \ref{tab:staged_losses} that $\eta(\embed(\xb), \treat)$ is not smooth enough, and the above GI loss helps. Note that the above loss is different from the gradient penalty used in other methods~\cite{senn, irm}, where the gradient norm $\Vert\frac{\partial \eta}{\partial t} \Vert$ is used as a regularizer.  Earlier work~\cite{gi_loss} has shown that such GI induced losses are more effective in increasing the smoothness of a deep network on a continuous input, rather than the proposals by~\citet{senn} and~\citet{irm}.

\subsection{Kernel Smoothed Inferred Counterfactual Outcomes}
To infer counterfactual responses for new treatments that are distant from the observed ones, we leverage the first assumption of Overlap: $P(\treatcf|\xb)>0$ for all $\xb$. On finite training data $D$, 
this implies that we need to rely on neighbors in $D$ from the $\embed(X)$ and $T$ space to infer counterfactual outcomes. However, two challenges arise: (1) combining proximity in both the high-dimensional $\embed(X)$ and low-dimensional $T$ space, and (2) unreliability of responses inferred from sparse neighborhoods.  We describe next how we handle these challenges via a Gaussian Process based estimator:

\paragraph{Gaussian Process for Estimating Counterfactual Response} Suppose we wish to infer $\yhat_i(\treatcf_i)$ for an observation $(\xb_i, y_i, \treat_i)$. % s.t. $|\treatcf_i - \treat_i| > \delta$. 
One option is to design a joint kernel over $\embed(X)$ and $T$, for example, the product kernel\cite{Bellot2022} or the Neural Tangent Kernel (NTK)~\cite{ntk} derived from the $\embed$, $\eta$ network.  However, we found the following two-stage approach with a few learned parameters to be more effective.

First we  account for proximity in $T$ space, by collecting the instances $(\xb_j,\treat_j,y_j)$, whose observed treatments $\treat_j$ are close to $\treatcf_i$, \ie, $|\treatcf_i-\treat_j|\le \gpdelta$, and define a nearest neighbor dataset $\NNDi$: 
\begin{align}
    \NNDi = \set{(\xb_j, \treat_j,  y_j) \in D  |\;\;  |\treatcf_i - \treat_j| \le \gpdelta}, \\ 
    \Xb_{\text{NN}},\yb_{\text{NN}} = [\xb_j \in \NNDi],[ y_j \in \NNDi] \label{eq:nnd}.
\end{align}

Then, to account for proximity in $\embed$ space, we fit a Gaussian Process (GP) using $\NNDi$ as an inducing set~\cite{titsias2009variational} to infer counterfactual responses. 
Specifically, we model $\y_i(\treatcf_i)$ as:
\begin{align*}
    & y_i  (\treatcf_i)  = f(\embed(\xb_i)) + \epsilon, \text{ where } \epsilon \sim \Ncal (0,\sigma^2) \quad \text{  with, } \\ & f(\embed(\xb_i)) \sim GP(0, K(\embed(\xb_i),\embed(\xb_i')))
\end{align*}
Then we estimate 
$\yhat_i (\treatcf_i)$, the mean of the posterior as:
\begin{align}
    \label{eq:mean_gp}
     \yhat_i (\treatcf_i)& %= \EE[y_i (\treatcf_i) \given \NNDi, \xb_i] 
      = K(\embed(\xb_i),\embed(\Xb_{\text{NN}})) V_{\text{NN}} ^{-1} \yb_{\text{NN}}. 
\end{align} 
where $V_{\text{NN}} =  [\sigma^2 \II+ K(\embed(\Xb_{\text{NN}}),\embed(\Xb_{\text{NN}}))]$. 
Further variance of the estimate $ \var[y_i (\treatcf_i)]$ %\given \NNDi,\xb_]$ 
is given by: 
\begin{align}
    \label{eq:var_gp}
&        K(\embed(\xb_i), \embed(\xb_i)) \nonumber\\
 &      -  K(\embed(\xb_i),\embed(\Xb_{\text{NN}})) V_{\text{NN}} ^{-1}  K(\embed(\xb_i),\embed(\Xb_{\text{NN}})).
\end{align}

\iffalse
This leads us to define the following counterfactual loss for the instance $(\xb_i,\treat_i,y_i)$ under $do(T) = \treatcf$, \ie,
\begin{align}
    \label{eq:gploss}
    &\ell(\eta(\embed(\xb_i),\treatcf),  \yhat_i (\treatcf))\nonumber \\
    &\qquad =  (\eta(\embed(\xb_i),\treatcf)-\yhat_i (\treatcf))^2
\end{align}
\fi
%
We leverage the GP variance to down-weigh unreliable outcomes and mitigate their impact on learned parameters. In particular, a large variance indicates a lack of nearby neighbors in $\NNDi$ for instance $\xb_i$. Thus, in a mini-batch setting, we first sample $\treatcf_i \sim P(T)$ for each individual in the batch, and then obtain mean and variance estimates using the GP for instances where $|\treatcf_i - \treat_i| > \delta$. We assign a weight $w_i(\treatcf_i) \propto \exp(-\var[y_i (\treatcf_i)])$ and 
 we apply a weighted loss to minimize the influence of unreliable counterfactuals on the overall loss.
\begin{align}
\gploss =   \sum_{\substack{i: \\ |\treatcf_i - \treat_i| > \delta}} 
 \dfrac{e^{-\var[y_i (\treatcf_i)]}} {\displaystyle\sum_{j} e^{-\var[y_j (\treatcf_j)]}} \ell(\eta(\embed(\xb_i),\treatcf_i),  \yhat_i (\treatcf_i))
\label{eq:gp-total}
\end{align}
We compute the GP quantities $\yhat_i(\treatcf_i), \var_i$ in Eq. \ref{eq:gp-total} inside \texttt{stop\_gradient}.
% \todo{We treat $\Phi$ as constant and only update the $\eta$ parameters.}
In practice, the GP is effective only when the outcomes inferred from it outperform those of a baseline model trained solely on the observational dataset. We show that the GP indeed produces better outcomes in Section ~\ref{sec:mot}. Moreover, we also show that the additional loss on sampled new treatments successfully addresses the discrepancy between the training and counterfactual test distributions, even after suppressing the impact of instances with high variance.

% \vspace{-4mm}
\subsection{Estimating Parameters}
\paragraph{Fixing GI+GP Parameters}
\label{sec:gigpparams}
The GI and GP layers for inferring the counterfactual outcomes require three parameters $\delta, \sigma$, and $\gpdelta$, which we fix based on the validation dataset. The estimation procedure does not involve training and thus is computationally efficient. The $\delta$ decides if we estimate $\yhat$ from the GI or the GP, and we fix it to minimize average L2 loss over the validation set.
We infer the GP-based mean and variance estimates on the validation dataset at observed treatments for different $\gpdelta$ and $\sigma$ values and compute the KS loss $\gploss$ on the validation dataset $D_{\text{val}}$ using Eqn \ref{eq:gp-total}. In particular, for a sample $(\xb, \treat, y)$ in the validation dataset, its loss is computed on the ground truth label $y$ for the observed treatment
$\treat$ as
$\frac{\exp(-\var[y (\treat)]) \cdot \ell(y,  \yhat (\treat))}{\sum_{(x', t', y') \in D_{\text{val}}} \exp(-\var[y' (\treat')])}$. Finally, we select the parameter values that yield the lowest loss.

\paragraph{Estimation of $\Phi,\eta$}
Our final objective function combines the three losses:
\begin{align} \label{eq:overall_obj}
    \min_{\embed, \eta}  ( \floss + \lambda_{\text{GI}} \giloss + \lambda_{\text{KS}} \gploss)
\end{align}
where $\lambda_{\text{GI}},  \lambda_{\text{KS}}$ are hyper-parameters that weigh the contributions of the individual loss terms. A brief description of the overall training and inference procedure is described in Algorithm~\ref{alg:giks}.

\begin{figure}[t]
 \begin{minipage}[t]{0.45\textwidth}
 \vspace{-2mm}
    \begin{algorithm}[H]
    \small
        \caption{\our\ training}
        \label{alg:giks}
        \begin{algorithmic}[1]
        \REQUIRE Training Data $D = \set{(\xb_i, \treat_i, y_i)}$, Validation Data $D_{\text{val}}, P(T), \lambda_{\text{GI}}, \lambda_{\text{KS}}$, \# of epochs $\mathrm{Epochs}$, starting epoch for GI loss $\mathrm{Epoch}_{\text{GI}}$, and for KS loss $\mathrm{Epoch}_{\text{GP}}$. 
        % \FOR{epoch $\in \set{1...E}$ }
        \STATE $\embed,\eta\gets $\textsc{Train}$(\text{VCNET}, D )$
                  \STATE $\delta, \sigma, \gpdelta \gets \textsc{FixGIGPParams}(D, D_{\text{val}},\Phi, \eta)$  
        %   \vspace{1mm}
          \FOR{$e \in  [\mathrm{Epochs}]$}
            \STATE $\mathrm{loss} \gets \textsc{FactualLoss}(D,\embed,\eta)$
             \STATE $\treatcf_i \sim P(T)$ for all $i \in [|D|]$  % $\treatcf_i \sim \mathrm{Unif}[0,1]$   for all $i\in [|D|]$
            \STATE $S_{\delta} \gets \set{(\xb_i, \treat_i, y_i): |t_i - \treatcf_i| < \delta}$
             \STATE $\mathrm{loss} \gets \mathrm{loss}+\textsc{GILoss}(S_{\delta} ,\set{\treatcf_i},\delta,\embed,\eta)$
                % \ELSE 
                \STATE $\mathrm{loss} \gets \mathrm{loss}+\textsc{KSLoss}(D\cp S_{\delta} ,\set{\treatcf_i},\gpdelta,\embed,\eta)$
            % \ENDIF 
            \STATE $\embed,  \eta\gets \textsc{GradDesc}(\mathrm{loss}  )$
        \ENDFOR
                \STATE \textbf{Return} $\embed, \eta$
\end{algorithmic}
\end{algorithm}
% \hrule
\end{minipage}
\begin{minipage}[t]{0.45\textwidth}
\small
\hrule
\begin{algorithmic}[1]
\STATE \textbf{function} {$\textsc{GILoss}(S_{\delta},\set{\treatcf_i},\delta,\embed,\eta)$}
             \FOR{$(\xb_i,t_i,y_i)\in S_{\delta}$}
                      \STATE $\yhat_i(\treatcf_i) \leftarrow y_i - (\treat_i -\treatcf_i) \frac{\partial \eta(\embed(\xb_i),\treat)}{\partial \treat}$  
            \ENDFOR
            \STATE $ \textbf{Return} \ \sum_{(\xb_i,t_i,y_i)\in   S_{\delta}}  \mathcal{L}(\eta(\embed(\xb_i), \treatcf_i), \yhat_i(\treatcf_i))$
     % \ENDFUNCTION
     \end{algorithmic}
\hrule
%%%
\begin{algorithmic}[1]
\STATE \textbf{function} {$\textsc{KSLoss}(D',\set{\treatcf_i},\gpdelta,\embed,\eta)$}
     \FOR{$(\xb_i,t_i,y_i)\in D'$}
        \STATE $\NNDi \leftarrow $\textsc{NNbr}$(D,\xb_i,\treat_i, \gpdelta)$ (Eq. \ref{eq:nnd})
        \STATE with stop\_gradient:
        \STATE \quad  $\text{m}_i \leftarrow \EE[y_i (\treatcf_i) \given \NNDi, \xb_i]$ (Eq. \ref{eq:mean_gp})
        \STATE \quad  $\hat{\sigma}^2 _i \leftarrow \text{Var}[y_i (\treatcf_i) \given \NNDi,\xb_i]$ (Eq. \ref{eq:var_gp})
    \ENDFOR
    \STATE Compute $\gploss$ using Eq.~\eqref{eq:gp-total}
    \STATE \textbf{Return}   $\gploss$  
    % \ENDFUNCTION
\end{algorithmic}
 \hrule
\end{minipage}
\end{figure}
 
\section{Related Work}
In this section, we briefly review the literature on both discrete and continuous treatment effect estimation.

\paragraph{Discrete Treatment Effect Estimation (DTE)}
Discrete treatment effect estimation can be categorized into three approaches: (1) {\em Sample re-weighting} methods~\cite{ipw_1, ipw_2}, which adjust counterfactual estimates using inverse  propensity scores but can be unstable without calibrated propensity score 
% We provide the pseudocode for training \our\ in Algorithm \ref{alg:giks}. 
models. (2) {\em Feature matching techniques} ~\cite{uri_learning_reps, matching_propensity, rubin1973matching, perfect_match, deepmatch}, which infer pseudo targets by aggregating labels of neighboring instances but are sensitive to distance metrics and lack reliability checks.
 (3) {\em Regularization-based} methods~\cite{cfrnet, dragonnet}, such as those using Integral Probability Metrics like Wasserstein distance and Maximum Mean Discrepancy, including Targeted Regularization. Such regularizers are introduced to improve Average Treatment Effect (ATE) estimation.
Other approaches include adversarial training methods \cite{ganite, weighted_factual_2}, variational autoencoders \cite{cevae, cevae_critical, reconsidering_gen}, and non-parametric Gaussian Process methods that discard the mean estimates and directly minimize the variance of counterfactual predictions \cite{bayesian_ite, overlapping_rep}.

\paragraph{Continuous Treatment Effect Estimation}
Existing literature on continuous treatment effect estimation has focused on two aspects:  (1) designing better neural architectures, and (2) designing better loss functions and regularizers.  The problem of CTE estimation was introduced in ~\cite{DRNet}, that proposed DRNet, that discretizes dosage values and uses separate last layers for each discrete dosage bin while sharing previous layers. VCNet ~\cite{vcnet}, on the other hand, avoids discretization by ensuring the smoothness of counterfactual predictions through a trainable spline function. Additionally, TransTEE ~\cite{TransTEE} proposed a Transformer-based representation network specifically designed for text datasets.

VCNet~\cite{vcnet} introduced a targeted regularizer to address the train-test mismatch and improve ATE estimation accuracy. In \cite{Bellot2022}, VCNet was extended to use the Hilbert Schmidt Independence criterion as a regularizer for generating treatment-independent embeddings.  Another way to enforce independence is by discretizing treatment groups and using an  IPM regularizer to make the representations of different treatment groups similar ~\cite{neurips22_2}. 
TransTEE ~\cite{TransTEE} further extended targeted regularizers to handle continuous treatments with a proposed probabilistic targeted regularizer. While targeted regularizers ensure consistent ATE estimates with asymptotic guarantees, they do not account for Individual Treatment Effect estimation. Another method that performs data augmentation like ours is SciGAN \cite{scigan}, which employs a generative adversarial network (GAN) to generate outcomes for new treatments. However, we will demonstrate the unstable training nature of this GAN-based approach in our experiments.

\begin{table*}[!ht]
    \centering
    \setlength\tabcolsep{4.0pt}
    \resizebox{0.8\textwidth}{!}{
    \begin{tabular}{l|r|r|r|r|r|r|r}
    \hline
        ~ &  TARNet & DRNet & SciGAN & TransTEE & VCNet+TR & VCNet+HSIC & \our\\ \hline\hline
		TCGA-0 &  1.673 (0.00) &  1.678 (0.00) &    2.744 (0.00) &  0.164 (0.25) &   \second{0.163 (0.31)} &  0.164 (0.29) &  \first{0.152} \\\hline
        TCGA-1 &  1.417 (0.00) &  1.465 (0.00) &  0.907 (0.00) &    0.146 (0.00) &  0.098 (0.03) &  \second{0.096 (0.08)} &  \first{0.080} \\\hline
        TCGA-2 &  3.365 (0.00) &  3.396 (0.00) &  1.359 (0.01) &    0.201 (0.00) &  0.152 (0.00) &  \second{0.144 (0.02)} &  \first{0.127} \\\hline
        IHDP   &  2.731 (0.00) &  3.068 (0.00) &      -- &    2.266 (0.00) &    2.263 (0.00) &  \second{1.961 (0.09)} &  \first{1.891} \\\hline
        NEWS   &  1.126 (0.00) &  1.163 (0.00) &      -- &    1.239 (0.00) &    1.107 (0.00) &    \second{1.104 (0.00)} &  \first{1.079} \\\hline

        \hline 
    \end{tabular}}
       \caption{Comparison of \our\ with baselines.
       % \ie, TARNet~\cite{cfrnet}, DRNet~\cite{DRNet}, SciGAN~\cite{scigan}, TransTEE~\cite{TransTEE}, VCNet+TR~\cite{vcnet}, and VCNet+HSIC~\cite{Bellot2022} 
       on CF error. The table includes the mean performance and within brackets, the $p$-values obtained from one-sided paired $t$-tests with \our\ as the base. Values less than 0.05 indicate statistically significant gains. "--" denotes models that did not converge.  
       % Green (Yellow) indicates the best (second best) performer. 
       \our\ outperforms all the baselines across all datasets, and the results are statistically significant for the majority of them.}\label{tab:final_res}
\end{table*}

{\renewcommand{\arraystretch}{1}%
\begin{table*}[!t]
    \centering
    \resizebox{0.82\textwidth}{!}{
    \begin{tabular}{l|r|r|r|r|r|r}
    \hline
        ~ & \multicolumn{2}{|c|}{TARNet} & \multicolumn{2}{|c|}{DRNet} & \multicolumn{2}{|c}{TransTEE} \\ \hline 
        ~ & Baseline & GIKS & Baseline & GIKS & Baseline & GIKS \\ \hline\hline
        TCGA-0 & 1.678 $\pm$ 0.027 & \first{1.077  $\pm$  0.034} & 1.673 $\pm$ 0.036 & \first{1.073  $\pm$  0.028} &  \first{0.164 $\pm$ 0.027} & 0.165  $\pm$  0.032 \\ \hline
        TCGA-1 & 1.465 $\pm$ 0.039  &  \first{0.555  $\pm$  0.011}  & 1.417 $\pm$ 0.049  & \first{0.556  $\pm$  0.013} & 0.146 $\pm$ 0.024  & \first{0.132  $\pm$  0.021}  \\ \hline
        TCGA-2 & 3.396 $\pm$ 0.059  & \first{0.746  $\pm$  0.045}  & 3.365 $\pm$ 0.079  & \first{0.747  $\pm$  0.041}  & 0.201 $\pm$ 0.011  & \first{0.172  $\pm$  0.179}  \\ \hline
        IHDP & 3.068 $\pm$ 0.126  &  \first{3.037 $\pm$ 0.227} & 2.731 $\pm$ 0.333  & \first{2.532 $\pm$ 0.169}  & 2.266 $\pm$ 0.182  & \first{2.023 $\pm$ 0.193}  \\ \hline
        NEWS & 1.163 $\pm$ 0.055  & \first{1.159 $\pm$ 0.088}  & \first{1.126 $\pm$ 0.069}  & 1.129 $\pm$ 0.135  & 1.239 $\pm$ 0.120  & \first{1.176 $\pm$ 0.179}   \\
        \hline 
    \end{tabular}}
    \caption{Performance comparison of \our\ with base model architecture adopted from three other state of the art methods, 
    \viz, TARNet,
     DRNet and
     TransTEE on the CF error. 
     % We measure the CF error on the test instances, that denotes the square root over the Mean Integral Squared Error assessed across all counterfactual treatments $\treatcf\in[0,1]$.
    }\label{tab:giks_architectures}
\end{table*}}

\section{Experiments}
\label{sec:expt}
\paragraph{Dataset} Following the prior work~\cite{vcnet, TransTEE, scigan}, we use five datasets, namely IHDP, NEWS, and TCGA(0-2) on three types of treatments. Across all datasets, $t\in [0,1]$. 
% They are summarized in Table 1 and further 
Details are deferred to the Appendix.

\paragraph{Methods}
\label{sec:sec:baselines}
We compare against six recent state-of-the-art baselines, \ie,  TARNet~\cite{cfrnet}, DRNet~\cite{DRNet}. SciGAN~\cite{scigan}, TransTEE~\cite{TransTEE} and VCNet+TR~\cite{vcnet}. Details of methods in Appendix.
% {\renewcommand{\arraystretch}{1.5}%
% \begin{table}[h]
%     % \vspace{-1mm}
%     \centering
%         \parbox{0.45\textwidth}{
%     \resizebox{0.45\textwidth}{!}{
%     \begin{tabular}{ c|r|r|r|r }
%     \hline
%         ~ & Search space &  TCGA(0-2) & IHDP & NEWS \\ \hline\hline
%         lrn rate & $[10^{-2}, 10^{-3}, 10^{-4}]$ &  $10^{-4}$ & $10^{-2}$ & $10^{-3}$ \\ \hline
%         $\lambda_{\text{GI}}$ & $[10^{-1}, 10^{-2}, 10^{-3}, 10^{-4}]$ &   $10^{-1}$ & $10^{-4}$ & $10^{-2}$ \\ \hline
%         $\lambda_{\text{KS}}$ & $[10^{-1}, 10^{-2}, 10^{-3}, 10^{-4}]$ &   $10^{-2}$ & $10^{-1}$ & $10^{-4}$ \\ \hline
%     \end{tabular}}
%      \caption{The hyperparameters. }  \label{tab:hpms}}
% \end{table}}

% {\renewcommand{\arraystretch}{1.5}%
\begin{table}[h]
    % \vspace{-1mm}
    \centering
        % \parbox{0.45\textwidth}{
    % \resizebox{0.45\textwidth}{!}{
    \begin{tabular}{ c|r|r|r }
    \hline
        ~ &  TCGA(0-2) & IHDP & NEWS \\ \hline\hline
        lrn rate &  $10^{-4}$ & $10^{-2}$ & $10^{-3}$ \\ \hline
        $\lambda_{\text{GI}}$ & $10^{-1}$ & $10^{-4}$ & $10^{-2}$ \\ \hline
        $\lambda_{\text{KS}}$ &  $10^{-2}$ & $10^{-1}$ & $10^{-4}$ \\ \hline
    \end{tabular}%}
     \caption{The hyperparameters. }  \label{tab:hpms}%}
\end{table}%}

\paragraph{Hyper-Parameter Estimation}
We allocate $30\%$ samples as validation dataset to tune hyperparameters.  Note, we depend only on \emph{factual error} and do not require counterfactual supervision even in the validation dataset.
\our\ has three hyperparameters: learning rate, $\lambda_{\text{GI}}$, $\lambda_{\text{GP}}$ that are optimized via grid search on factual error of the validation dataset. Further, the GP employs a cosine kernel. We use a batch size of 128, the AdamW optimizer, and early stopping based on factual error on the validation dataset. The results of hyperparameter tuning are presented in  Table \ref{tab:hpms}.

% Case study on Algorithmic Recourse

\paragraph{Evaluation Metric (CF Error)}
\label{sec:expts}
Following existing literature, we evaluate performance using CF Error, short for counterfactual estimation error, that measures the prediction accuracy for arbitrary treatments applied on test instances, thus making it suitable for the ICTE problem.
%\paragraph{CF Error (Mean Integral Squared Error)} 
Given $N$ test instances, we define the CF error as: $\sqrt{\frac{1}{N} \textstyle  \sum_{i= 1} ^{N} \int_{\treatcf_i = 0} ^{1} (y_i(\treatcf_i) - \yhat_i(\treatcf_i))^2 P(\treatcf_i) d\treatcf_i}$. 
% The probabilistic integral inside the square root is indicative of the performance being measured on arbitrary treatments sampled from $P(T)$ applied to test instances.
The error integrates over treatments sampled from $P(T)$ applied to the test instances.
In practice, since it is difficult to determine the test time treatment distribution $P(T)$, our default is using a uniform distribution, as followed in earlier work \cite{scigan}, and we present ablations on other candidates of $P(T)$.

\subsection{Comparison with SOTA Methods}
 Table~\ref{tab:final_res} compares our method against all state-of-the-art CTE methods (Section~\ref{sec:sec:baselines}) based on CF Error.  We make the following observations:
 %
 % \begin{enumerate*}
     % \item
     \textbf{(1)} \our\ consistently outperforms all baselines on CF error across all datasets, highlighting the suitability of our loss function for ICTE estimation. The statistically significant gains in performance, as indicated by the computed $p$-values from a one-sided paired $t$-test with \our\ as the base, further support the superiority of our approach, except for the TCGA-0 dataset where performance is comparable to the next competing baselines.
    \textbf{(2)} SciGAN, despite incorporating instance-level counterfactual losses like our approach, demonstrates poor performance due to the challenges associated with training the min-max objectives in adversarial training. Our experiments revealed instances where the error significantly increased for specific dataset seeds, resulting in high result variance. The lack of a control mechanism akin to our GP variance to filter unreliable counterfactual supervision prevented model convergence when the counterfactual supervision provided by the generator was flawed.
     \textbf{(3)} DRNet and TARNet suffer from poor performance due to their discretization of treatments, which leads to their $\eta$ network being less sensitive to changes in $\treat$. 
   \textbf{(4)} VCNet with two regularizers: Targeted Regularizer and HSIC are both worse than \our.
     Although HSIC is a better regularizer for the ICTE problem, it still operates at a distribution level rather than at an instance level like \our.
%    
     % \textbf{(5)} 

 % \end{enumerate*}

 % \vspace{0.3cm}

\if{0}
\begin{table}
    % \vspace{-4mm}
    \centering
         \parbox{0.3\textwidth}{
        \resizebox{0.3\textwidth}{!}{
        \begin{tabular}{c|r|r}
        \hline
        ~ & TransTEE & VCNet \\ \hline\hline
        \multirow{ 1}{*}{TCGA-0}  & 1.3 $\times 10^{-1}$ & 5.6 $\times 10^{-2}$ \\ \hline
        \multirow{ 1}{*}{TCGA-1}  & 4.4$\times 10^{-5}$ & 2.0 $\times 10^{-1}$ \\ \hline
        \multirow{ 1}{*}{TCGA-2} & 1.4$\times 10^{-8}$ & 3.6 $\times 10^{-3}$\\ \hline
        \multirow{ 1}{*}{IHDP} & 8.5$\times 10^{-10}$ & 3.6$\times 10^{-8}$\\ \hline
        \multirow{ 1}{*}{NEWS} & 6.6 $\times 10^{-18}$ & 1.6 $\times 10^{-4}$ \\  
        \hline
        \end{tabular}}
     \caption{$p$-values.\label{tab:p_test_res}}}
\end{table}

Next, we assess the statistical significance of the improvements of \our\ over the state-of-the-art methods, and the results are presented in Table~\ref{tab:final_res}.  We conduct one-sided paired $t$-tests with our method as the reference. The results, shown in Table~\ref{tab:p_test_res}, indicate that the improvements achieved by \our\ are statistically significant for all datasets except TCGA-1. \todo{p-values will change. I will bring vcnet+hsic also}

\fi

\begin{figure*}
    \begin{minipage}[t]{\textwidth}
      % \centering
      \begin{minipage}{0.47\textwidth}
          \begin{figure}[H]
            \centering \includegraphics[width=\textwidth]{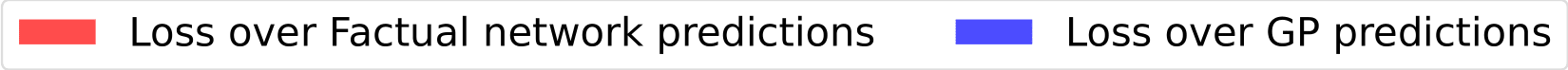} \\
              \subfloat[\centering IHDP ]{{\includegraphics[width=0.49\textwidth]{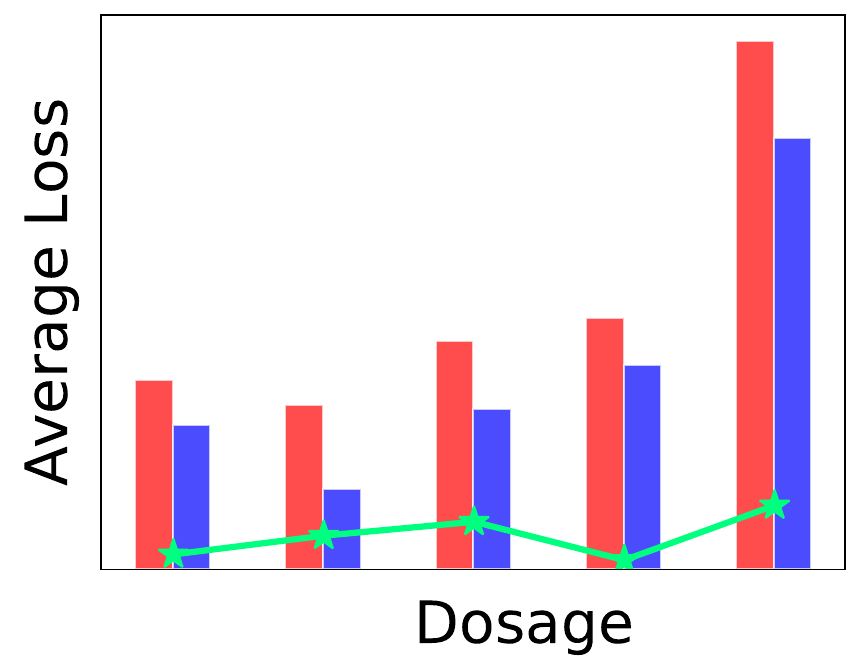} }}%
            \subfloat[\centering  Synthetic]{{\includegraphics[width=0.49\textwidth]{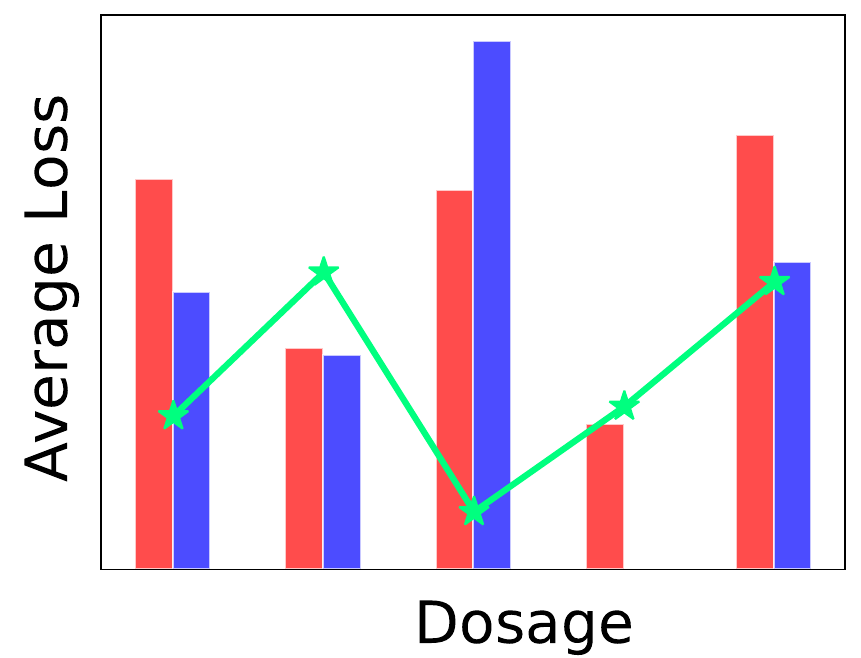} }}%
               \caption{Losses on Counterfactuals}
                \label{fig:ctr_losses}
          \end{figure}
      \end{minipage} \hfill 
      \begin{minipage}{0.48\textwidth}
          \begin{figure}[H]
              \centering \includegraphics[width=0.9\textwidth]{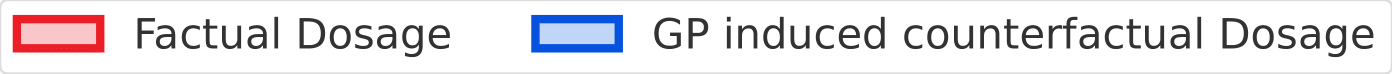} \\
              \subfloat[\centering IHDP ]{{\includegraphics[width=0.49\textwidth]{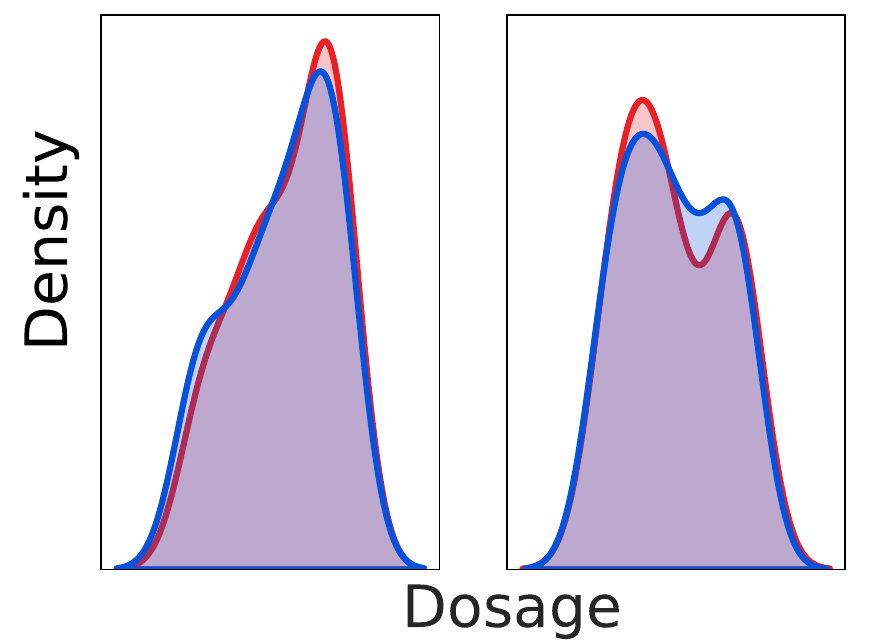} }}%
            % \qquad
            \subfloat[\centering  Synthetic]{{\includegraphics[width=0.49\textwidth]{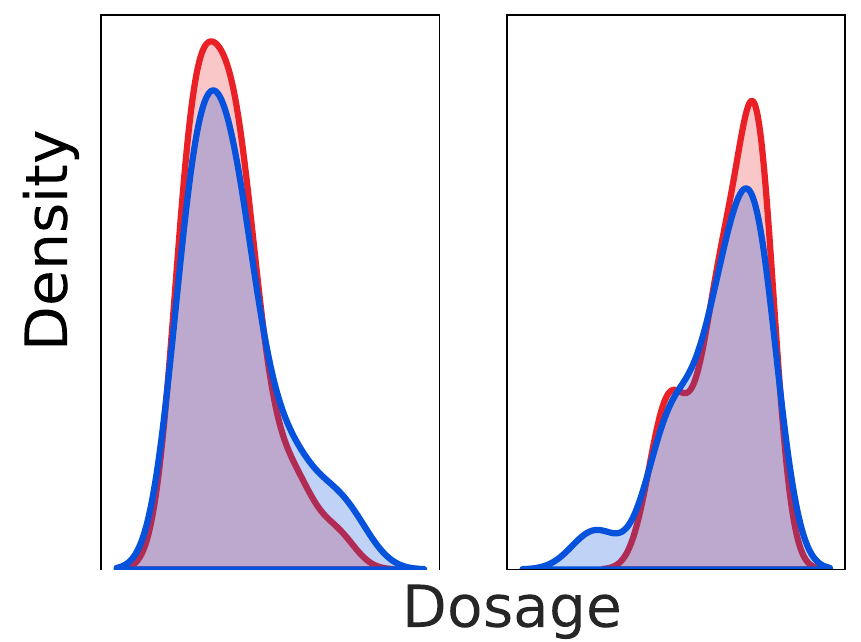} }}%
               \caption{Training Dosage distribution}
                \label{fig:ctr_dist}
          \end{figure}
      \end{minipage}
\end{minipage}
\end{figure*}

\subsection{\our\ with Different Base Architectures} \label{sec:different_base}
 The model-agnostic nature of \our\ allows it to be integrated with various base architectures. While VCNet is chosen for its effectiveness with continuous treatments, we explore the potential of \our\ with other networks such as TARNet, DRNet, and TransTEE next. Our experiments, presented in  Table \ref{tab:giks_architectures}, show that \our\ enhances the performance of all the base networks.

\subsection{Why GIKS Works?} \label{sec:mot}
To see why our approach produces better counterfactual estimates, we 
conduct experiments using the IHDP dataset, and a synthetic dataset taken from ~\citet{vcnet}, to answer the following questions:
\begin{itemize}[leftmargin=*]
    \item Are inferred counterfactuals using neighbors in $\embed$ and $T$ space more accurate than those obtained from a baseline model that is trained solely on observational dataset $\trnD$? 
    \item For a given individual $\xb$, do our augmented loss help reduce the divergence $D(P(\xb, \treat) || P(\xb)P(\treat))$ between training and test distribution?
\end{itemize}

We answer the first question by contrasting the performances of two models: (1) a baseline factual model trained solely on the observational dataset $\trnD$ using factual loss $\floss$, and (2) our \our\ model trained using algorithm \ref{alg:giks}. Now, for \our\ to work, we need GP to produce counterfactual estimates that are more accurate than the factual model. To assess this, we compare the losses of the factual model with the losses of GP estimates obtained using Eq. \ref{eq:mean_gp} for training instances at randomly sampled new treatments. The results in Figure \ref{fig:ctr_losses} demonstrate that GP produces more accurate counterfactual estimates than the baseline's $\eta$ network, except for the middle bin $(0.4, 0.6]$ in the synthetic dataset, which has limited instances. The line plot in Figure \ref{fig:ctr_losses} shows the distribution of examples in each bin. By performing a one-sided paired $t$-test, we confirm that GP's losses are statistically significantly lower than the factual losses with a $p$-value of 0 for both datasets, providing further confidence in GP's higher accuracy.

To answer the second question, we compare the treatment distributions used to train the factual model and \our\ for different individuals in the dataset. Figure \ref{fig:ctr_dist} presents the results for both datasets, showcasing the treatments at which losses were imposed during training for two arbitrarily chosen candidates and their 30 nearest neighbors. We observe that \our\ reduces the skew in the treatment distribution, leading to a lower divergence $D(P(\xb, \treat) || P(\xb)P(\treat))$. To further validate this observation, we compute the HSIC metric, which resulted in divergence values of 8.30 (0.94) for the factual model of synthetic (IHDP) and 4.9 (0.63) for \our\ of Synthetic (IHDP) dataset. These results, along with the previous findings, provide insights into the effectiveness of \our\ for counterfactual inference.

\subsection{Ablation Study}
% Here, we study the sensitivity of \our\ to changes in its various components.

\paragraph{Impact of the Three Losses on \our\ Performance}

\begin{figure}
  \begin{center}
    \includegraphics[width=0.23\textwidth]{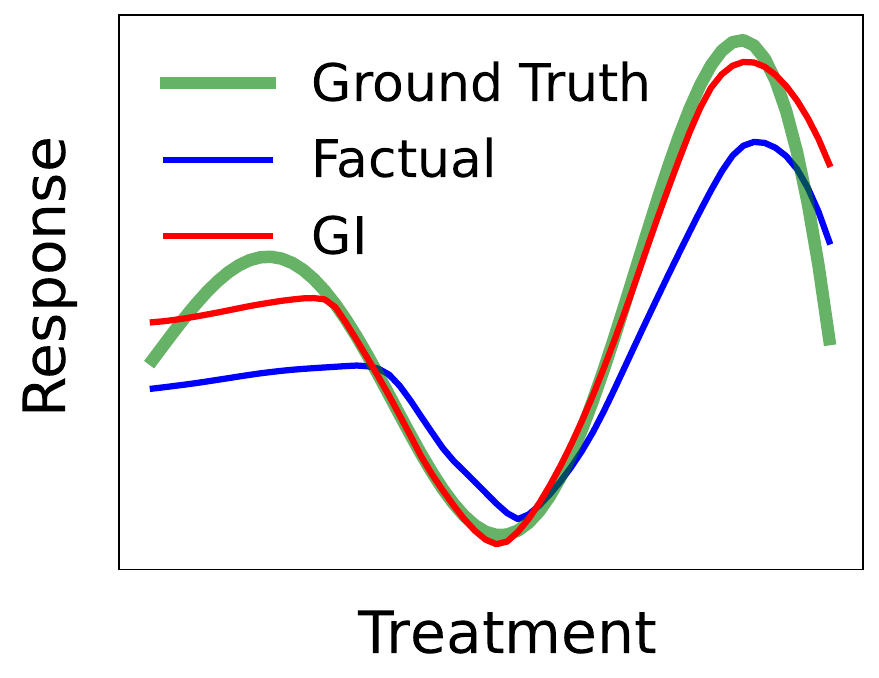} 
    \includegraphics[width=0.23\textwidth]{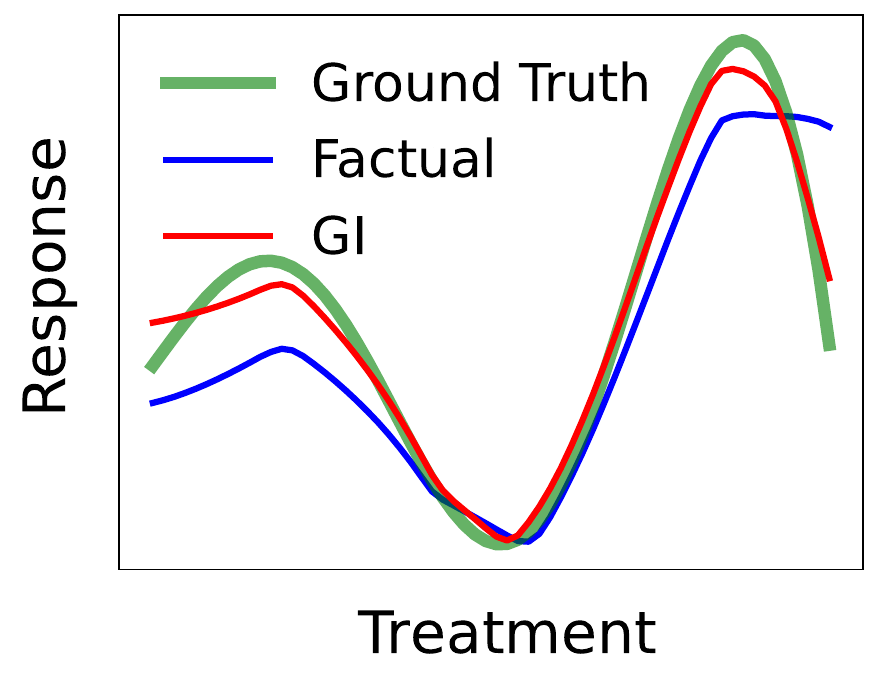} 
  \end{center}
  \caption{IHDP Individualised Dose-Response Function}
  \label{fig:ihdp_smooth}
\end{figure}

{\renewcommand{\arraystretch}{1.3}%
\begin{table}
    % \vspace{-2mm}
    \centering
        % \parbox{0.45\textwidth}{
        \resizebox{0.45\textwidth}{!}{
            \begin{tabular}{l|r|r|r|r}
                \hline 
                    Dataset & $\fctloss$ & $  + \lambda_{\text{GI}} \giloss$ &   $\lambda_{\text{KS}} \gploss$ & \our
                    % \vtop{\hbox{\strut $\fctloss + \lambda_{\text{GI}} \giloss$}\hbox{\strut $+\lambda_{\text{KS}} \gploss$}}  
                    \\ \hline\hline
                    TCGA-0 &  0.18 (0.10) &  0.17 (0.26) &   0.17 (0.21) &  \first{0.15} \\\hline
                    TCGA-1 &   0.09 (0.08) &  0.09 (0.62) &  0.09 (0.63) &  \first{0.09} \\\hline
                    TCGA-2 &  0.17 (0.03) &   0.17 (0.06) &  0.16 (0.01) &  \first{0.12} \\\hline
                    IHDP   &  2.05 (0.00) &  1.91 (0.32) &   1.96 (0.10) &  \first{1.89} \\\hline
                    NEWS   &  1.09 (0.00) &   1.08 (0.41) &   1.08 (0.23) &  \first{1.07} \\
                    \hline
            \end{tabular}}
     \caption{CF error for the models trained using different combinations of our loss components.}\label{tab:staged_losses}%}
\end{table}}

In this experiment, we analyze the impact of each of the three proposed losses in \our\ by measuring the CF error achieved for models that are trained on different combinations of \our\ losses until convergence. The results are summarized in Table \ref{tab:staged_losses}.   First, observe that neither GI, nor GP is superior over the other, and this lets us conclude that both our loss components have a non-trivial effect on the performance of \our.  Second, the GI loss alone manages to provide gains over VCNet and complements the smoothness provided by the spline parameterization of the VCNet model. While VCNet focuses on smooth parameter variations of the $\eta$ network with $t$, our $\giloss$ ensures that the predictions of the $\eta$ network also vary smoothly with $t$. These two methods work together to achieve a smooth ADRF at an instance level.  We observe that $\giloss$ helps in smoothing the predicted Dose-Response Function for treatments close to the observed treatments, as elucidated in Figure \ref{fig:ihdp_smooth} for two training instances.

\begin{table}
    % \vspace{-4mm}
    \centering
        % \resizebox{0.45\textwidth}{!}{
        \begin{tabular}{l|r}
    \hline
        ~ & CF error \\ \hline\hline
        NTK & 1.925 (0.35) \\ \hline
        Dot product & 1.882 (0.43)  \\ \hline
        cosine & \first{1.842}  \\ \hline
    \end{tabular}%}
     \caption{\our\ performance using different kernels over $5$ seeds of IHDP dataset.}\label{tab:gp_kernel}
\end{table}

\paragraph{Kernel Exploration}
We study the impact of different types of kernels on the performance of \our. We experimented with three kernels: cosine kernel $K(\embed(\xb),\embed(\xb')) =  \frac{\embed(\xb)^\top \embed( \xb')}{||\embed( \xb)|| ||\embed( \xb)||}$, dot product kernel 
$K(\embed(\xb),\embed(\xb')) =  \embed(\xb)^\top \embed( \xb')$, and finally NTK kernel $ 
K(\embed(\xb_i),\embed(\xb_j)) = {\nabla_{\eta, \embed} \eta(\embed(\xb_i), t_i)}^\top\nabla_{\eta, \embed} \eta(\embed(\xb_j), t_j)$. 
Table~\ref{tab:gp_kernel} summarizes the results for the IHDP dataset averaged over $5$ seeds. We present the $p$-values of one-sided paired $t$-test with cosine kernel as the base within bracket. We observe that: 
(1) Cosine kernel performs better than Dot-product kernel, perhaps 
because Cosine is invariant to the scale of the embeddings. (2) Even though the NTK kernel defines a joint kernel over $\Xcal$ and $\Tcal$ space, unlike \our\ that used a two-stage kernel, computing the NTK Kernel is both computation and memory intensive. %Additionally, the table shows large $p$-values, indicating comparable CF errors across different kernels. 

\paragraph{Impact of Alternative $P(T)$}
We experimented with alternative distributions for sampling new treatments during training: (1){\em  Marginal treatment distribution}, where we sample treatments from the empirical marginal treatment distribution in the observational dataset.
(2){\em  Inverse Propensity (IP) distribution}, where we discretized the treatments into $10$  bins and trained a propensity model $\pi: \Xcal \rightarrow [10]$ that predicts the treatment bin. Then we sample treatments from a bin chosen with a probability $P(\bullet \given \xb) = \frac{1}{\pi(\bullet \given \xb)}$. 
To make the results comparable, Table \ref{tab:pt_sampling} presents the CF error tested on uniform distribution across these three training strategies. We also report $p$-values of the one-sided paired $t$-test with default sampling. We observe that \our\ performance is similar across all the sampling strategies, that is possibly because (1) For the IHDP and news dataset,  their marginal treatment distribution is close to uniform (3) For IP sampling, the propensity model achieved an accuracy of $100\%$ in predicting the correct bin for both datasets. The inverted propensity scores gave negligible weight to the predicted bin and approximately equal weight to the other bins leading to close to uniform sampling.

\begin{table}
    % \vspace{-4mm}
    \centering
        % \resizebox{0.3\textwidth}{!}{
            \begin{tabular}{l|r|r}
                \hline 
                ~ 	& IHDP 	& NEWS \\
                \hline \hline
                Marginal &  1.902  (0.42) &  1.086  (0.09) \\\hline
                IP &   1.903  (0.41) &   1.080  (0.36) \\\hline
                Uniform &           \first{1.891} &           \first{1.079} \\
                \hline 
            \end{tabular}%}
     \caption{Sampling strategies for $P(T)$. }\label{tab:pt_sampling}
\end{table}

\if{0}
\paragraph{Effect of $\delta$ neighborhood in KS}
In this experiment, we vary $\delta$ and check its impact on the inducing points selection for GP.  We vary from $\delta$ as small as $0.025$ to $0.1$ on the IHDP dataset for 5 seeds and record the test performance in Table \ref{tab:delta_perf}. A larger $\delta$ brings in a lot of noisy covariates and a smaller $\delta$ results in a paucity of instances in $\NND$ for the GP to smoothen on. We observe $\delta = 0.05$ to work well in this experiment.

{\renewcommand{\arraystretch}{1.5}%
\begin{table}[H]
    \caption{Impact of $\delta$ on inducing points set of GP for $5$ seeds on IHDP dataset. We observe that $\delta = 0.05$ works the best.\label{tab:delta_perf}}
    \centering
    \resizebox{\columnwidth}{!}{
    \begin{tabular}{c|r|r|r|r}
    \hline\hline
        ~ & $\delta=0.025$ & $\delta=0.050$ & $\delta=0.075$ & $\delta=0.100$ \\ \hline\hline
        CF error & 1.912$\pm$0.211 & 1.860$\pm$0.138 & 1.914$\pm$0.194 & 1.916$\pm$0.161 \\ \hline
        Factual error & 5.075$\pm$1.702 & 4.378$\pm$0.780 & 4.730$\pm$1.400 & 4.699$\pm$0.760  \\ \hline\hline
        %NEWS MISE & 0.051$\pm$0.015 & 0.046$\pm$0.019 & 0.054$\pm$0.010 & 0.058$\pm$0.018 \\ \hline
        %NEWS Factual error & 1.382$\pm$0.101 & 1.408$\pm$0.085 & 1.381$\pm$0.112 & 1.382$\pm$0.112  \\ \hline
    \end{tabular}}
    
\end{table}}
\fi
{\renewcommand{\arraystretch}{1.5}%
\begin{table}[!h]
    \centering
    \setlength\tabcolsep{3.2pt}
    \resizebox{0.48\textwidth}{!}{
    \begin{tabular}{l|r|r|r|r}
    \hline\hline
        ~ &  TransTEE & VC+TR & VC+HSIC & GIKS \\  \hline\hline
	CF Error & \first{0.42 $\pm$ 0.01} & 0.43 $\pm$ 0.01 & 0.43 $\pm$ 0.01 & \first{0.42 $\pm$ 0.01} \\ \hline
        Policy Err. & 0.22 $\pm$ 0.03 &	0.19 $\pm$ 0.02 & 0.18 $\pm$ 0.00 & \first{0.17 $\pm$ 0.01} \\	\hline
        Rec. Acc. & 0.59 $\pm$ 0.02 & 0.60 $\pm$ 0.02	& 0.62 $\pm$ 0.04 &  \first{0.63 $\pm$ 0.01} \\
        \hline \hline
    \end{tabular}}
        % \vspace{0.2cm}
       \caption{Performance of \our\ vs. baselines on recourse for image setting for skin lesion diagnosis. We report mean $\pm$ std. deviation averaged over $5$ seeds. }
       \label{tab:case_study}
\end{table}}
\subsection{Case Study}
We conducted two case studies illustrating the application of treatment effect estimation to Algorithmic Recourse where the goal is to predict the treatment that yields the best outcome. We present the skin lesion diagnosis experiment in this section and defer the insulin prediction experiment to regulate glucose levels to the appendix.

\paragraph{Algorithmic Recourse for Mobile Skin Lesion Diagnosis} We consider an application where users submit skin images to a diagnostic classifier designed for lesion detection. In the event of a user uploading a low-quality image, the classifier might yield an inaccurate diagnosis. Our objective is to guide users in adjusting their image capture settings to enhance diagnostic accuracy. We employed a commonly used skin lesion detection dataset, featuring seven labels, with the classifier predicting a distribution across these seven categories. Our goal in algorithmic recourse is to recommend image settings for which the classifier confidence --- probability of the predicted class --- is high.
As part of the setting (treatment), we consider brightness level of the image. For training the model for counterfactual inference we created an observation dataset $D$, where each instance is a 3-tuple $(\treat_i, \phi(\xb_i), \rho_i)$. Here, $\xb_i$ is an initial image, $\phi(\xb_i)$ 
 is its representation, $\treat_i$ denotes 
 the treatment applied to produce post-treatment image $\xb' _i $, and $\rho_i$  
 reflects the classifier's confidence on the treated image. 
 We introduced selection bias by making the 
 observed treatments depend on $\phi(\xb_i)$.
Testing the classifier on a test set with treatments sampled uniformly from $[-0.5, 0.5]$, we observed an accuracy of 44\%. %This motivates our goal of learning the ICTE with brightness as the treatment variable and classifier confidence as the response. Our approach involves
We evaluate the performance on VCNet architecture of baselines vs. GIKS using Recourse Accuracy which is defined as the accuracy achieved on the test dataset after treating the test images with predicted optimal dosages.  We report the recourse performance comparing \our\ with other competing baselines in Table \ref{tab:case_study}.   We observed that GIKS achieves the highest recourse accuracy of $63 \pm 0.02\%$. We also report dosage policy error which is the difference between the accuracy achievable at the best brightness setting, and the accuracy at the recommended setting by the trained model.  Even on this metric, our model performs better.

\iffalse{0}

{\renewcommand{\arraystretch}{1.5}%
\begin{table}[!h]
    \centering
    \setlength\tabcolsep{3.0pt}
    \resizebox{0.47\textwidth}{!}{
    \begin{tabular}{l|r|r|r|r|r}
    \hline\hline
        ~ & TransTEE & TransT(GIKS) & VCNet+TR & VC+HSIC & GIKS \\  \hline\hline
	CF Error & 0.65 (0.22) & 0.64 (0.35) & 0.63 (0.48) &  0.63 (0.50) & \first{0.62} \\ \hline
        Policy Error & \first{1.39 (0.63)} & \first{1.39 (0.63)} & 1.85 (0.21) & 1.59 (0.43) & 1.52 \\	\hline
    \end{tabular}}
        \vspace{0.2cm}
       \caption{Performance of \our vs. baselines for insulin treatment on VCNet. We averaged over $5$ seeds. }
       \label{tab:insulin}
\end{table}}

\paragraph{Recommending Insulin Treatments}
Next, we use the counterfactual inference task to recommend insulin dosages to patients based on their monitored blood glucose levels two hours before a meal and the amount of carbohydrates consumed in the meal.  We generate the dataset for this task using the Simglucose simulator~\cite{Simglucose2018} a Python implementation of the FDA-approved UVa/Padova Simulator~\cite{UVA2018} for research purposes only. The simulator includes 30 virtual patients, 10 adolescents, 10 adults, and 10 children.  More details of our experiments are described in the appendix.  Table~\ref{tab:insulin} presents the dosage policy error of different methods on the VCNet architecture, and we observe that GIKS provides best results.

\fi

\section{Conclusion} 
We addressed estimating Individualized Continuous Treatment Effects from observational data, where treatments are confounded with covariates. Our method aims to reduce the mismatch between observed data distribution and the independence needed for counterfactual estimation by sampling new treatments for training instances. 
% The complexity of inferring counterfactual responses depends on the proximity between observed and new treatments. 
We devised two strategies for synthesizing pseudo targets and applying instance-specific counterfactual losses. Experiments on benchmark ICTE datasets showed statistically significant gains over other baselines. We also presented experimental results on two potential medical applications. Future work could include a more thorough investigation of these applications and extending our approach to cases where the overlap assumption is violated as recently highlighted in ~\cite{neurips22_1}.

\bibliography{GIKS-main}

\newpage
\appendix
\onecolumn
\begin{center}
    \Large { \bf Appendix \\
    \normalsize (Continuous Treatment Effect Estimation Using Gradient Interpolation and Kernel Smoothing)
    }
\end{center}

\section{Sim Glucose Dataset}
In this section, we conduct experiments using a real-world glucose simulator to demonstrate the effectiveness of our approach in predicting the impact of insulin on the risk of low and high glucose levels in individuals with Type-1 Diabetes. We used an FDA-approved UVa/Padova Simulator (2008 version) to simulate the dose-response function $\mu$. Our goal is to predict the risks of pumping different dosages of insulin into patients with Type-1 diabetes and thereby design a control algorithm that avoids high risks. 
% The observational dataset for this task is generated using an open-source implementation of this simulator.

The simulator models risk as a function of the following inputs obtained from a patient: (a) patient covariates, a $13$ dimensional vector that uniquely identifies each patient, (b) the amount of insulin pumped in the patient, (c) the amount of carbs consumed by the patient is her meal. The simulator first predicts the blood glucose levels which are then converted to Magni risk using the formula: 

\begin{equation}
\text{risk}(b) = 10 \cdot (c_0 \cdot \log(b)^{c_1} - c_2)^2
\end{equation}

where $b$ denotes the predicted blood glucose levels. Here, $c_0$ and $c_1$ are constants adjusted to assign lower risks to blood glucose values within the range $[70, 180]$ and higher risks otherwise as a safe blood glucose limit is assumed to be in the range of $b\in [70, 180]$. Therefore, the objective of any insulin control algorithm is to administer the appropriate insulin doses to maintain blood glucose levels within this safe range. Following prior work \cite{fox2019reinforcement}, each entry in our observational dataset consists of the covariates $\xb$ captured as the measured blood glucose levels 2 hours prior to taking a meal, and the number of carbs consumed during the meal. Now, given a treatment $t$, that denotes the amount of insulin pumped into the blood, our response $y$ captures the maximum risk value over the next two hours of taking the meal.

In our main submission, we did not report results on the TransTEE baseline, which we present here:

{\renewcommand{\arraystretch}{1.8}%
\begin{table}[!h]
    \centering
    \setlength\tabcolsep{3.0pt}
    \resizebox{0.6\textwidth}{!}{
    \begin{tabular}{l|r|r|r|r|r}
    \hline\hline
        ~ & TransTEE & TransTEE (GIKS) & VCNet+TR & VCNet+HSIC & GIKS \\  \hline\hline
	CF Error & 0.658 (0.223) & 0.643 (0.350) & 0.628 (0.482) &  0.627 (0.500) & \first{0.626} \\ \hline
        Dosage Policy Error & \first{1.391 (0.628)} & \first{1.391 (0.628)} & 1.856 (0.209) & 1.591 (0.427) & 1.519 \\	\hline
    \end{tabular}}
        \vspace{0.2cm}
       \caption{Performance of \our vs. baselines for insulin treatment on VCNet. We averaged over $5$ seeds. }
       \label{tab:insulin}
\end{table}}

We observe that for this task, TransTEE architecture performs well on the Dosage policy Error. So, we tested the effectiveness of applying the \our\ losses on the TransTEE architecture and observed that \our\ maintains the performance on the policy error while improving the CF error.

\xhdr{Results on Skin-lesion diagnosis case study}
We report the results on the TransTEE baseline for this case study also here:

{\renewcommand{\arraystretch}{1.5}%
\begin{table}[!h]
    \centering
    \setlength\tabcolsep{3.2pt}
    \resizebox{0.6\textwidth}{!}{
    \begin{tabular}{l|r|r|r|r}
    \hline\hline
        ~ &  TransTEE & VCNet+TR & VCNet+HSIC & GIKS \\  \hline\hline
	CF Error & \first{0.42 $\pm$ 0.01} & 0.43 $\pm$ 0.01 & 0.43 $\pm$ 0.01 & \first{0.42 $\pm$ 0.01} \\ \hline
        Dosage Policy Error & 0.22 $\pm$ 0.03 &	0.19 $\pm$ 0.02 & 0.18 $\pm$ 0.00 & \first{0.17 $\pm$ 0.01} \\	\hline
        Recourse Accuracy & 0.59 $\pm$ 0.02 & 0.60 $\pm$ 0.02	& 0.62 $\pm$ 0.04 &  \first{0.63 $\pm$ 0.01} \\
        \hline \hline
    \end{tabular}}
        \vspace{0.2cm}
       \caption{Performance of \our vs. baselines on recourse for image setting for skin lesion diagnosis. We report mean $\pm$ std. deviation averaged over $5$ seeds. }
       \label{tab:case_study}
\end{table}}

In this experiment, we found that GIKS applied to the VCNet architecture performed the best in all the assessed metrics. We reported std. deviation here as the performance of \our\ was found to be non-overlapping with the baselines.

\section{Code / Datasets}
We have uploaded the code for \our\ along with the Supplementary material. 
% Also we have released the anonymized version of the code in the URL: \url{https://anonymous.4open.science/r/giks-5EF3/README.md}.

\section{Limitations}
% \todo[L]{Profs review}
A limitation of our approach is that we make the differentiability (A3) and overlap (A1) assumptions in the dataset without an explicit verification of whether they hold. However, one safe guard we implemented was to depend on the validation dataset for determining the contribution of additional losses arising out of these losses via the GP parameters and the weights assigned to the counterfactual losses ($\lambda_{\text{GI}}, \lambda_{\text{KS}}$). In the worst case, we will not provide any gains beyond the baseline factual model. A second limitation is the running time associated with inferring counterfactual responses. While inferring GI based responses is cheap, inferring  GP based responses requires a nearest neighbor search over the entire training dataset.

\section{Datasets Description}
Here we discuss more details on the datasets that we used in our work.

\paragraph{TCGA (0--2)  \cite{scigan}} 
The TCGA dataset, obtained from The Cancer Genome Atlas project, consists of data on various types of cancer in $9659$ individuals. Each individual is characterized by $4000$ dimensions of gene expression covariates. These covariates are log-normalized and further normalized to have unit variance. The treatment variable represents the dosage of the drug taken by the patient, while the synthetic response models the risk of cancer recurrence. In our experiments, we use three versions of the TCGA dataset proposed in \cite{scigan}, referred to as TCGA(0), TCGA(1), and TCGA(2).

\paragraph{IHDP  \cite{ihdp}}  was originally collected from the Infant Health Development Program and used for binary treatment effect estimation. In the dataset, treatments were assigned through a randomized experiment. It consists of $747$ subjects with $25$ covariates. For the continuous treatment effect (CTE) problem, the dataset was adapted in \cite{vcnet} by assigning synthetic treatments and targets.

\paragraph{News \cite{news}}  was initially a binary treatment effect dataset but was adapted in \cite{scigan} for continuous treatments and targets. The treatment variable in this dataset represents the amount of time a user spends reading a news article, while the synthetic response aims to resemble user satisfaction. The dataset consists of $3000$ randomly sampled articles from the New York Times, with $2858$ bag-of-words covariates. 

For each of the datasets, we generate several versions of it by using different seeds following prior work. The dataset statistics were provided in Table 1 in the main paper.

\section{Synthetic Treatment and Response Generation}
We adopt the dataset generation process from the prior methods and we {\em quote} the details of dataset generation here for completeness. \cite{DRNet, vcnet, scigan}.

\paragraph{IHDP}
For treatments in $[0,1]$, we generate responses as in \cite{vcnet}. Specifically, for a given $\xb$, we generate $y$ and $t$ as follows: 
\begin{align}
    \tilde{t}\given \xb &=\frac{2x_1}{1+x_2}+\frac{2\max(x_3,x_5,x_6)}{0.2+\min(x_3,x_5,x_6)} +2\tanh\left(\frac{5\sum_{i\in S_{\text{dis},2}}(x_i-c_2)}{|S_{\text{dis},2}|}-4+\mathcal{N}(0,0.25)\right), \\
    & t=(1+\exp(-\tilde{t}))^{-1} \\
    y\given \xb, t & =\frac{\sin(3\pi t)}{1.2-t}\cdot\tanh\left(\frac{5\sum_{i\in S_{\text{dis},1}}(x_i-c_1)}{|S_{\text{dis},1}|}\right) + \frac{\exp(0.2(x_1-x_6))}{0.5+5\min(x_2,x_3,x_5)}\big) + \mathcal{N}(0,0.25)
\end{align}

where
$S_{\text{con}}=\{1,2,3,5,6\}$ is the index set of continuous covariates, $S_{\text{dis},1}=\{4,7,8,9,10,11,12,13,14,15\}$, $S_{\text{dis},2}=\{16,17,18,19,20,21,22,23,24,25\}$ and $S_{\text{dis},1}\bigcup S_{\text{dis},2}=[25]-S_{\text{con}}$. 

Further $c_1=\mathbb{E}\left[\frac{\sum_{i\in S_{\text{dis},1}} x_i}{|S_{\text{dis},1}|}\right]$, $c_2=\mathbb{E}\left[\frac{\sum_{i\in S_{\text{dis},2}} x_i}{|S_{\text{dis},2}|}\right]$. 

\paragraph{NEWS} First, we  generate $v_1', v_2', v_3'$ from $\mathcal{N}(0,1)$, and set $v_i= \frac{v_i'}{\| v_i' \|}$
\begin{align}
    t\given \xb &= \text{Beta}\left(2, \left\| \frac{v_3^{\top}}{2v_2^{\top}\xb} \right\|\right),
    y'\given \xb, t = \exp\left(\frac{v_2^{\top}\xb}{v_3^{\top}\xb}-0.3\right) \\
    y\given \xb,t &= 2(\max(-2,\min(2,y'))+ 20v_1^{\top}\xb)  (4(t-0.5)^2 + \sin\left(\frac{\pi}{2}t\right) + \mathcal{N}(0,0.5) \label{eq:news_response}
\end{align}

\paragraph{TCGA(0-2)}
First generate $v_1', v_2', v_3'$ from $\mathcal{N}(0,1)$, and set $v_i= \frac{v_i'}{\| v_i' \|}$
Add noise $\epsilon \sim \mathcal{N}(0,0.2)$. Then, dosage $d\given \xb,t \sim \text{Beta}(\embed, t)$, where $\embed$ (default as 2) is the dosage selection bias. $t_t = \frac{\embed-1}{d^*} + 2-\embed$, with $d^*$ as optimal dosage for that treatment.

For TCGA (0), we generate $y$ and $d^*$
as    $y\given \xb,d = 10(v_1^{\top}\xb + 12d v_3^{\top}\xb  - 12 d^2 v_3^{\top}\xb ),
    d^* = \frac{v_2^{\top}\xb}{2v_3^{\top}\xb}$.

For TCGA(1), we generate $y$ and $d^*$
as follows:    $y\given\xb,d = 10((v_1)^{\top}\xb + \sin(\pi (\frac{v_2^{\top}\xb}{v_3^{\top}\xb}d))),
    d^* = \frac{v_3^{\top}\xb}{2 v_2^{\top}\xb}$.

For  TCGA(2),   $ y\given \xb,d = 10(v_1^{\top}\xb + 12d(d-0.75\frac{v_2^{\top}\xb}{v_3^{\top}\xb})^2),
    d^* = 0.25\frac{v_2^{\top}\xb}{v_3^{\top}\xb} \text{ if } \frac{v_2^{\top}\xb}{v_3^{\top}\xb}\ge 1, \text{ else } 1$.

\section{Additional performance metrics}

We present results on four additional metrics that are also reported in the counterfactual inference literature for assessing Average Treatment Effect, finding the dosage that yields the best outcome, etc. We report the mean and one standard deviation from it obtained across different seeds of the datasets.

\paragraph{Factual Error (RMSE)} 
Here, we report the root mean squared error between the gold factual targets in the test data and the predicted outcomes as
\begin{align}
    \label{eq:ferror}
    \sqrt{\frac{1}{N}\textstyle \sum_{i= 1} ^{N} (y_i(\treat_i) - \widehat{y_i}(\treat_i))^2}.
\end{align}
This metric evaluates the predictive performance of the models on test instances using only their observed dosages. 

{\renewcommand{\arraystretch}{1.5}%
\begin{table*}[!ht]
 \caption{C
   Comparison of \our\ with baselines TARNet~\cite{cfrnet}, DRNet~\cite{DRNet}, SciGAN~\cite{scigan}, TransTEE~\cite{TransTEE}, VCNet+TR~\cite{vcnet}, and VCNet+HSIC~\cite{Bellot2022} on the Factual error assessed on the test dataset. Factual Error represents the predictive performance specifically on the observed treatments applied to test instances. However, it is not a representative metric for the ICTE problem. `--' indicates non-convergence.}
    \label{tab:final_res_fact}
    \centering
    \setlength\tabcolsep{3.0pt}
    \resizebox{0.9\textwidth}{!}{
    \begin{tabular}{l|r|r|r|r|r|r|r}
    \hline\hline
        ~ &  TARNet & DRNet & SciGAN & TransTEE & VCNet+TR & VCNet+HSIC & \our\\ \hline\hline
        TCGA-0 &  1.173 $\pm$ 0.177 &   1.189 $\pm$ 0.16 &  3.796 $\pm$ 0.719 &     \first{0.2 $\pm$ 0.02} &  0.219 $\pm$ 0.006 &  \second{0.205 $\pm$ 0.036} &  0.217 $\pm$ 0.003 \\ \hline
        TCGA-1 &  0.887 $\pm$ 0.025 &   0.92 $\pm$ 0.034 &  2.384 $\pm$ 0.283 &  \first{0.199 $\pm$ 0.019} &  0.215 $\pm$ 0.005 &  \second{0.205 $\pm$ 0.036} &  0.219 $\pm$ 0.005 \\ \hline
        TCGA-2 &  2.041 $\pm$ 0.047 &   2.073 $\pm$ 0.07 &  2.757 $\pm$ 1.352 &  0.205 $\pm$ 0.024 &  \first{0.227 $\pm$ 0.004} &  0.239 $\pm$ 0.006 &  \first{0.227 $\pm$ 0.005} \\ \hline
        IHDP   &  3.398 $\pm$ 0.289 &  2.978 $\pm$ 0.383 &      -- &  2.323 $\pm$ 0.257 &   2.55 $\pm$ 0.264 &  2.192 $\pm$ 0.337 &   2.22 $\pm$ 0.321 \\ \hline
        NEWS   &  1.207 $\pm$ 0.049 &  1.202 $\pm$ 0.042 &      -- &  1.356 $\pm$ 0.057 &  \second{1.327 $\pm$ 0.047} &   \second{1.14 $\pm$ 0.055} &  1.197 $\pm$ 0.039 \\ 
        \hline \hline
    \end{tabular}}
   
\end{table*}}

Table \ref{tab:final_res_fact} shows the results on factual error, where we  observe a slight increase on some datasets. This is expected because our counterfactual losses prevent the overfitting of models to just the factual distribution using imposing losses on uniformly sampled counterfactuals with synthesized pseudo-targets. In summary, \our\ trades off some factual accuracy to effectively handle instances that are more likely to occur during the inference stage. TransTEE performs better for factual error because they use powerful transformers to fit the factual distribution.

\paragraph{Dosage Policy Error (DPE)} It is defined as
    \begin{align}
        \frac{1}{N} \sum_{i= 1} ^{N} (y_i(\treat^*) - y_i(\widehat{\treat}))^2   
    \end{align}   

 where $\treat^*$ is the ground truth treatment for the best possible outcome, and $\widehat{\treat}$ is the predicted best treatment. This metric is more relevant in medical datasets than in TCGA. The results are shown in Table \ref{tab:dpe}.

{\renewcommand{\arraystretch}{1.5}%
 \begin{table*}[!ht]
    \centering
    \resizebox{0.7\textwidth}{!}{
    \begin{tabular}{l|l|l|l|l|l}
    \hline
        ~ & TARNet & DRNet & TransTEE & VCNet & \our\\ \hline\hline
        \multirow{ 1}{*}{TCGA(0)} & 0.402 $\pm$ 0.193 & 0.523 $\pm$ 0.414 & 0.120 $\pm$ 0.075 & \second{0.117 $\pm$ 0.095} & \first{0.073 $\pm$ 0.037}\\ \hline
        \multirow{ 1}{*}{TCGA(1)} & 0.446 $\pm$ 0.028 & 0.452 $\pm$ 0.023 & {0.009 $\pm$ 0.007} & \second{0.006 $\pm$ 0.001} & \first{0.004 $\pm$ 0.005} \\ \hline
        \multirow{ 1}{*}{TCGA(2)} & 0.793 $\pm$ 0.063 & 0.819 $\pm$ 0.057 & 0.007 $\pm$ 0.007 & \second{0.005 $\pm$ 0.006} & \first{0.004 $\pm$ 0.007}  \\ \hline
    \end{tabular}}
    \caption{ This table shows the Dosage Policy Error (DPE) for the TCGA(0-2) datasets. We see that \our\ outperforms the baselines for TCGA(0) and TCGA(2). for TCGA(1), we VCNet gives us the best performance in terms of DPE. These results show that counterfactual losses imposed by \our\ are indeed effective in determining the optimal dosages specific to each individual that can lie beyond the training distribution.}  \label{tab:dpe}
\end{table*}}

\paragraph{Average Mean Squared Error (AMSE)} This metric is defined as
 \begin{align}
     \frac{1}{N}  \sum_{i= 1} ^{N} \int_{0} ^{1} (y_i(\treatcf) - \widehat{y_i}(\treatcf))^2 P(\treatcf) d\treatcf
 \end{align}
     
 It assesses the error in the predicted responses integrated over treatments sampled from training propensity distribution $\pi$, and averaged across all the individuals in test data. The results are shown in the Table \ref{tab:amse}.

\section{AMSE Results}
\begin{figure}[h]%
    \centering
    \subfloat[\centering IHDP ]{{\includegraphics[width=0.22\textwidth]{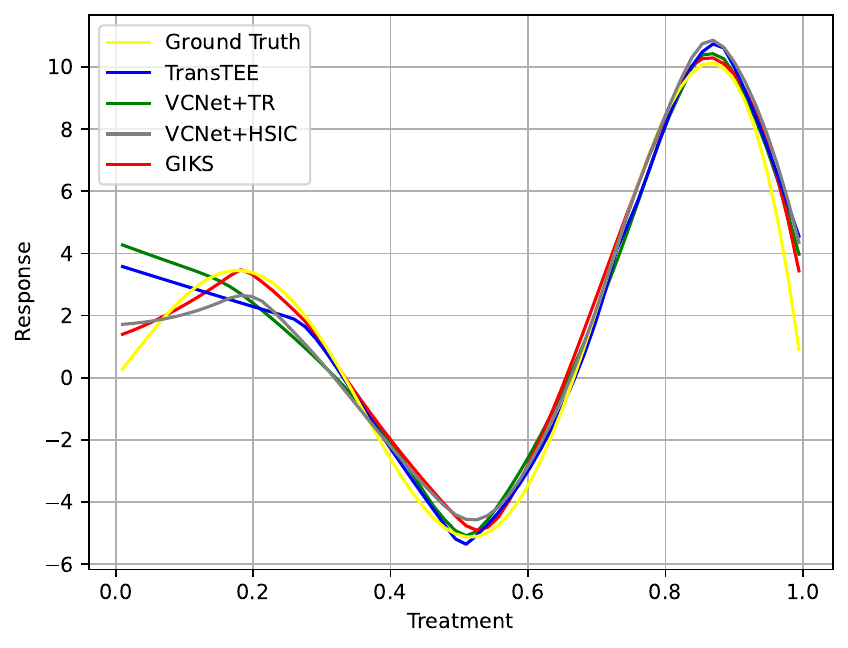} }}%
    \centering
    \subfloat[\centering News ]{{\includegraphics[width=0.22\textwidth]{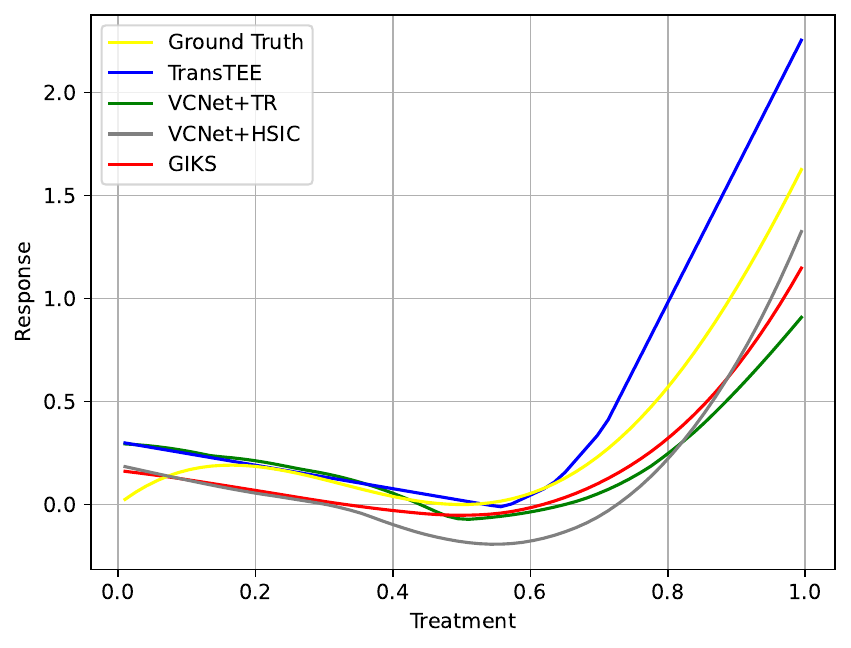} }}%
    
    \centering
    \subfloat[\centering TCGA-0 ]{{\includegraphics[width=0.22\textwidth]{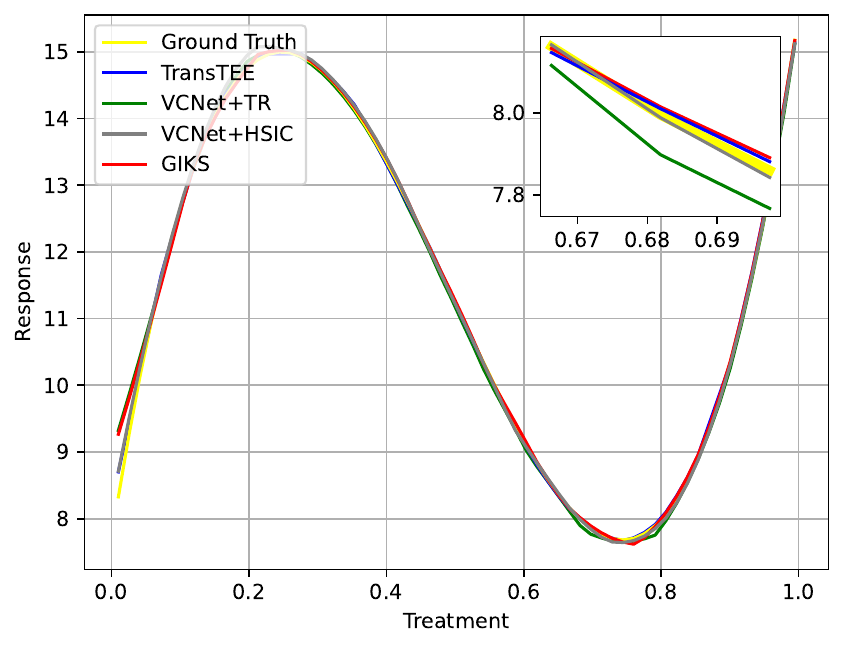} }}%  
    \centering
    \subfloat[\centering TCGA-1 ]{{\includegraphics[width=0.22\textwidth]{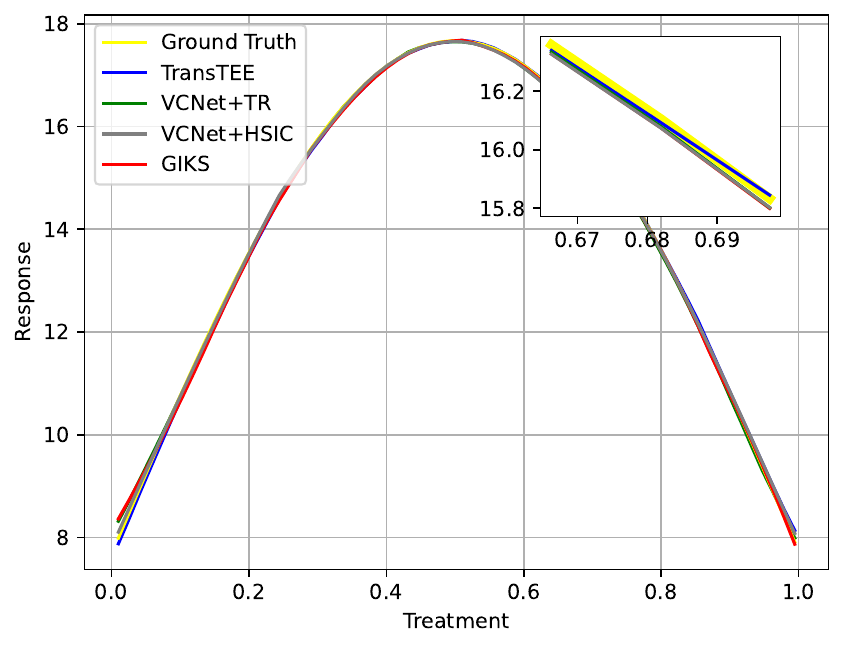} }}%
    \centering
    \subfloat[\centering TCGA-2 ]{{\includegraphics[width=0.22\textwidth]{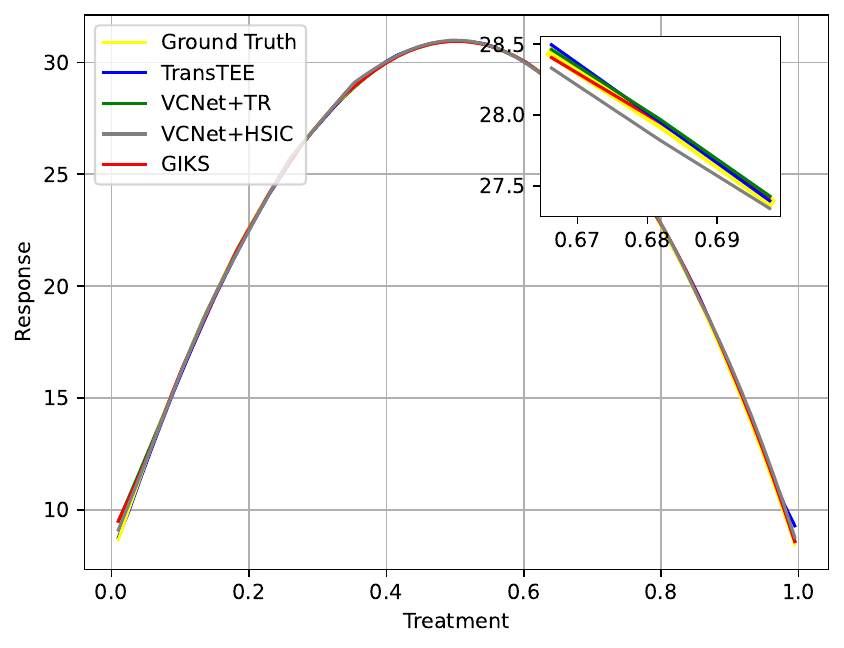} }}%
    \vspace{0.2cm}
    \caption{This Figure shows the ADRF curves for all the baselines and \our.}
\end{figure}

{\renewcommand{\arraystretch}{1.5}%
\begin{table*}[!ht]
    \centering
    \resizebox{0.7\textwidth}{!}{
    \begin{tabular}{l|l|l|l|l|l}
    \hline
        ~ & TARNet & DRNet & TransTEE & VCNet & \our\\ \hline\hline
        \multirow{ 1}{*}{TCGA(0)} & 5.422 $\pm$ 0.131 & 5.407 $\pm$ 0.183 & \second{0.0294 $\pm$ 0.009} & 0.039$\pm$0.012 & \first{0.028$\pm$0.0174}\\ \hline
        \multirow{ 1}{*}{TCGA(1)} & 2.778$\pm$0.110 & 2.637$\pm$0.159 & {0.0267 $\pm$ 0.009} & \second{0.012$\pm$0.003} & \first{0.009$\pm$0.001} \\ \hline
        \multirow{ 1}{*}{TCGA(2)} & 12.586$\pm$0.416 & 12.364$\pm$0.604 & 0.0498 $\pm$ 0.007 & \second{0.030$\pm$0.007} & \first{0.024$\pm$0.006}  \\ \hline
        \multirow{ 1}{*}{IHDP} & 3.477 $\pm$ 1.435 & 6.660 $\pm$ 0.460 & 0.951 $\pm$ 0.426 & \second{0.883 $\pm$ 0.487} & \first{0.519 $\pm$ 0.380}  \\ \hline
        \multirow{ 1}{*}{NEWS} & 0.195 $\pm$ 0.028 & 0.200 $\pm$ 0.026 & \first{0.018 $\pm$ 0.012} & 0.048 $\pm$ 0.021 & \second{0.034 $\pm$ 0.020} \\ \hline
    \end{tabular}}
    \caption{ This table shows the Average Mean Squared Error (AMSE) values meant for Average Treatment Effect assessment for baselines and \our. We observe the trend of these results also to be in alignment with the trend for CF Error. Except for NEWS, \our\ outperforms the baselines on all datasets. }\label{tab:amse}
\end{table*}}

We present the AMSE results for all the datasets in the table \ref{tab:amse}. For TCGA(0-2) we present the DPE in the table \ref{tab:dpe}.  We find that on AMSE, our results are consistent with the MISE metric reported in the main paper.  On DPE we are the best on two out of three cases.

\section{ADRF curves}
\label{sec:adrf}
 In the main paper, we showed the Dose Response Function for an individual on IHDP and TCGA(0) datasets. Here we show the {\em average} dose Response Function (ADRF) obtained for all the datasets in the Figure \ref{fig:ihdp_smooth}. For each of these curves, we obtain the ARDF function by averaging the responses of $30$ nearest neighbors in the $\Phi$ space.

For the NEWS dataset, the methods are not able to fit the ADRF curve perfectly because the training instances have an irreducible additive gaussian noise sampled from $\mathcal{N}(0,0.5)$ in their response $y$, as shown in the Eqn \ref{eq:news_response}. This has not been an issue with other datasets since the scale of $y$ for them is much larger compared to $0.5$; while for NEWS, it is comparable. This leads to the irreducible noise in the dataset having a significant impact on the response variable and this makes the ground truth ADRF curve more irregular.

All curves predict the treatment effect very well for TCGA(0-2), hence the curves seem coincident. We provide a zoomed-in section of the graph for a small range of treatments to better illustrate that \our\ achieves a marginally better fit of the ground-truth ADRF compared to the baselines.

\iffalse{0}
\section{Case Study: Algorithmic Recourse}

Here, we show the impact of the treatment (brightness) on the skin-lesion dataset that we considered in the Algorithmic Recourse case study.

\begin{figure}[h]%
    \centering
    \subfloat[\centering $t = -0.50$ ]{{\includegraphics[width=0.22\textwidth]{images/medical/ISIC_0025069-0.5.png} }}%
    \centering
    \subfloat[\centering $t=-0.375$ ]{{\includegraphics[width=0.22\textwidth]{images/medical/ISIC_0025069-0.625.png} }}%
    \centering
    \subfloat[\centering $t=-0.250$ ]{{\includegraphics[width=0.22\textwidth]{images/medical/ISIC_0025069-0.75.png} }}%
    \centering
    \subfloat[\centering $t=0.000$ ]{{\includegraphics[width=0.22\textwidth]{images/medical/ISIC_0025069-1.0.png} }}%
    \\
    \centering
    \subfloat[\centering $t=+0.125$ ]{{\includegraphics[width=0.22\textwidth]{images/medical/ISIC_0025069-1.125.png} }}%
    \centering
    \subfloat[\centering $t=+0.250$ ]{{\includegraphics[width=0.22\textwidth]{images/medical/ISIC_0025069-1.25.png} }}%
    \centering
    \subfloat[\centering $t=+0.375$ ]{{\includegraphics[width=0.22\textwidth]{images/medical/ISIC_0025069-1.375.png} }}%
    \centering
    \subfloat[\centering $t=+0.500$ ]{{\includegraphics[width=0.22\textwidth]{images/medical/ISIC_0025069-1.4375.png} }}%
    \vspace{0.2cm}
    \caption{This Figure shows the impact of applying the treatment to a test image. The image in panel (d) shows the original image in the skin-lesion dataset.}
\end{figure}
\fi

\section{Theoretical Analysis} \label{sec:theory}
The starting point for our theoretical analysis is Theorem 1 from \citet{Bellot2022} that gives the generalization bound for the counterfactual prediction error. We \textbf{quote} the theorem here:

\textbf{Theorem 1:} (Generalization bound for the average counterfactual error). Assume that the unit expected loss functions $l_{h, \phi}(x, t) / B_\phi \in G$ for $B_\phi>0$ for all $x \in \mathcal{X}$ and $t \in \mathcal{T}$, where $G$ is a family of functions $g: \mathcal{R} \times \mathcal{T} \rightarrow \mathbb{R}$. Then,
\begin{equation}
\label{eq:thm}
    \epsilon_{C F} \leq \epsilon_F+B_\phi \cdot \sup _{g \in G}\left|\int_{\mathcal{T}} \int_{\mathcal{R}} g(\mathbf{r}, t) \cdot\left(p_\phi(\mathbf{r}) p_\phi(t)-p_\phi(\mathbf{r}, t)\right) d \mathbf{r} d t\right|    
\end{equation}
In this expression, $\epsilon_{CF}$ represents the error observed on samples from the counterfactual test distribution, while $\epsilon_F$ corresponds to the error on the training distribution. The key insight derived from this theorem is that the additional error during testing is bounded by the second term, which quantifies the dependence between $X$ and $T$. This term equals zero only when $X$ is independent of $T$. Consequently, reducing the counterfactual error is possible by augmenting samples that meet two criteria: (a) the augmented samples provide reliable supervision for $y$, and (b) the augmented dataset exhibits less dependence between $X$ and $T$ than the given training dataset.

In this section, we explore a simplified scenario where both $X$ and $T$ are real-valued ($X\in \RR$ and $T\in \RR$). The joint distribution between $X, T$ during test is $\mathcal{N}(0, \sigma^2 I)$ where $I$ is the $2\times 2$ identity matrix, where as during training $X, T$ are correlated $\mathcal{N}(0, \begin{pmatrix}
  \sigma^2 & \rho\\ 
  \rho & \sigma^2
\end{pmatrix})$ where $\rho$ controls the covariance between $X, T$. We assume that the potential outcome $\mu(x, \treat)$ is $\tau'$-smooth in $\xb$ and $\delta'$-smooth in $t$. Additionally, for an observed sample $(x_i, t_i, y_i)$, we aim to synthesize a pseudo-outcome $\hat{y}(x_i, \treatcf_i)$ for a new treatment $\treatcf_i \ne t_i$. Let us define the data augmentation with \our\ as successful if the error introduced in the pseudo-outcome is bounded by $\tau$, expressed as $|\hat{y}(x_i, \treatcf_i) - \mu(x_i, \treatcf_i)| \leq \tau$. We analyze the specific conditions under which \our\ can achieve this objective.

Because $\mu(x, t)$ is $\tau'$ smooth in $x$, we have:
\begin{align}
    & \frac{|\mu(x, t) - \mu(x', t)|}{|x-x'|}  < \tau'
    \implies |\mu(x, t) - \mu(x', t)| < \tau' |x-x'|
\end{align}

By setting $ \tau' |x-x'| < \tau$, we establish the condition $|x-x'| < \frac{\tau}{\tau'}$. This condition indicates that if \our\ interpolates the pseudo-outcome $\hat{y}(x_i, \treatcf_i)$ from samples ${(x_j, t_j, y_j)}$ in the training dataset, ensuring $t_j = \treatcf_i$ and $|x - x_j| < \frac{\tau}{\tau'}$, then the augmentation effectively meets the error tolerance criteria.

A similar analysis, using $\delta'$ smoothness of $\mu$ with respect to the treatment, demonstrates that if \our\ interpolates the pseudo-outcomes using samples where $|\treatcf_i - t_j| < \frac{\tau}{\delta'}$, then the augmentation is successful.

Thus if the nearest neighbor dataset $\NNDi$ used for synthesizing pseudo-outcomes is obtained as $\{(x_j, t_j, y_j) \text{ s.t. } |x - x_j| < \frac{\tau}{2\tau'} \text{ and } |\treatcf_i - t'| < \frac{\tau}{2 \delta'}\}$, and if this set is non-empty, we can consider the augmentation successful. Note that, in our original algorithm, we do not filter instances in $\NNDi$ based on covariate distance, but we do drop high variance instances while training which can be thought of as dropping instances where success criterion is not met.

Next, we compute the probability with which the sample $(x_i,\treatcf_i)$ will be added to the augmented set.  This requires us to find at least one $(x_j,t_j)$ in $D$ that satisfies the proximity condition on $x_i$ and $\treatcf_i$.  The probability of this event can computed as:
\begin{align}
 P_A(\treatcf_i|x_i) \propto   1 -  \left( 1 - \int_{x \in [x_i - \frac{\tau}{2\tau'}, x_i + \frac{\tau}{2\tau'}]} \int_{t \in [\treatcf_i - \frac{\tau}{2\delta'}, \treatcf_i + \frac{\tau}{2\delta'}]} \mathcal{N}( [x \;\; t]; 0, \begin{pmatrix}
  \sigma^2 & \rho\\ 
  \rho & \sigma^2 
\end{pmatrix}) dt\; dx \right)^N
\end{align}

and this we approximate as:

\begin{align}
 P_A(\treatcf|x_i) \propto&   \min\left(1, N\int_{x \in [x_i - \frac{\tau}{2\tau'}, x_i + \frac{\tau}{2\tau'}]} \int_{t \in [\treatcf_i - \frac{\tau}{2\delta'}, \treatcf_i + \frac{\tau}{2\delta'}]} \mathcal{N}( [x \;\; t]; 0, \begin{pmatrix}
  \sigma^2 & \rho\\ 
  \rho & \sigma^2
\end{pmatrix}) dt\; dx\right) 
\end{align}

\begin{figure}[h]
    \centering
    \includegraphics[width=0.7\textwidth]{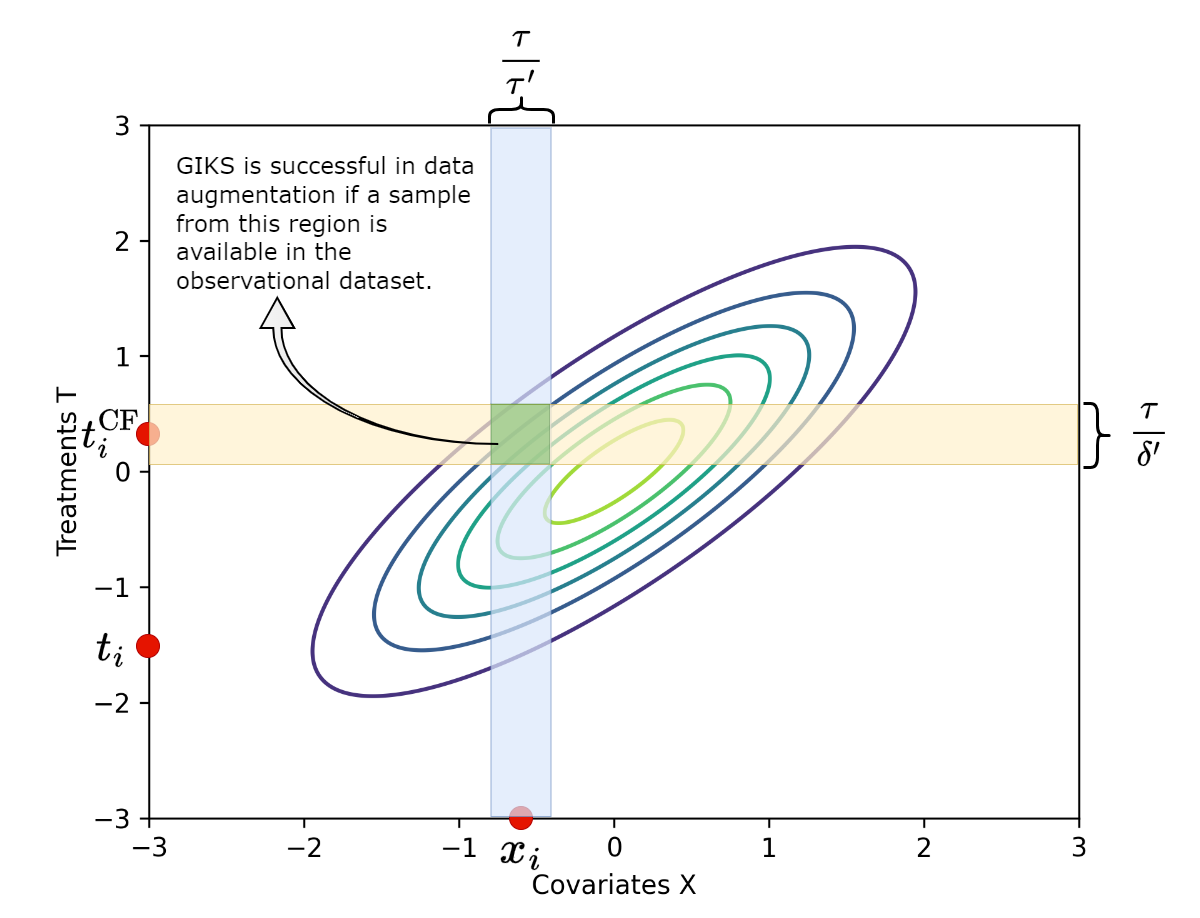}
    \caption{This figure depicts the analysis described in Section \ref{sec:theory}. In the training dataset, the covariates X and treatments $T$ are jointly Gaussian with mean 0 and covariance $\begin{pmatrix}
  1 & 0.8 \\ 
  0.8 & 1
\end{pmatrix}$. The goal is to synthesize an outcome $\hat{y}(x_i, \treatcf_i)$ such that the error in the synthesis is less than $\tau$. \our\ achieves this when the underlying observational dataset has at least one sample from the shaded green region in the middle. The shaded blue (yellow) regions in the figure represent the band of width $\frac{\tau}{ \tau'}$ ($\frac{\tau}{\delta'}$) around $x_i$ ($\treatcf_i$) that are considered in $\NNDi$.}
    \label{fig:gauss_mot}
\end{figure}

By overlap assumption (A1), the above probability is non-zero and  as the size of the training dataset increases, $P_A(\treatcf_i|x_i)$ will be independent of $\treatcf_i$ for most $x_i$, causing the treatment distribution to be independent of $X$ in the augmented data.  In contrast, in the training set $P(T|x_i)$ could be peaked if $\rho$ is large.
We have depicted a pictorial overview of this analysis in the Figure \ref{fig:gauss_mot}.  

We are working on deriving an expression for the distributional distance in Equation~\ref{eq:thm} for the above case and extending it to the general case.  

\subsection{\our\ addresses distribution mismatch}
In this section, we aim to explain the performance of \our\ observed in the experiments by showing that the augmented samples effectively reduce the distribution mismatch between the training and test datasets while controlling the CF Error.
To assess this, we compute the dependence between the covariate and treatment distributions ($D(P(\xb, T) || p(\xb)p(T))$) using the HSIC metric ~\cite{Bellot2022}, a kernel-based statistical independence test. The results are summarized in Table \ref{tab:hsic_obs}, where higher HSIC values indicate a stronger dependence between the $X$ and $T$ variables in the dataset.

Notably, our analysis reveals that the HSIC metric of the dataset augmented with our method is lower than that of the original dataset. This observation reinforces the effectiveness of our approach in mitigating the distribution mismatch between the training and test datasets. We also provide the corresponding $p$-values in brackets for the factual case with GIKS counterpart as the baseline, indicating statistically significant reductions in HSIC for three out of the five baseline cases. For TCGA-1 and TCGA-2, we observed large $p$-values because the factual model already exhibits strong performance in terms of the CF error, leaving limited room for improvement through our augmentation.

{\renewcommand{\arraystretch}{1.5}%
\begin{table*}[!t]
    \centering
    \resizebox{0.7\textwidth}{!}{
    \begin{tabular}{l|r|r|r|r}
    \hline
        ~ & \multicolumn{2}{|c|}{CF Error} & \multicolumn{2}{|c}{HSIC metric} \\ \hline 
        ~ & \makecell{VCNet on \\Observational Data} & \makecell{VCNet on \\Augmented Data} & \makecell{VCNet on \\Observational Data} & \makecell{VCNet on \\Augmented Data} \\ \hline\hline
        TCGA-0 &        0.19 (0.11) &             \first{0.15} &    2.13 (0.04) &         \first{0.57} \\
        TCGA-1 &         0.1 (0.08) &             \first{0.09} &    0.43 (0.27) &         \first{0.40} \\
        TCGA-2 &        0.17 (0.02) &             \first{0.13} &    0.41 (0.16) &         \first{0.37} \\
        IHDP &         2.06 (0.0) &             \first{1.89} &     0.72 (0.0) &         \first{0.34} \\
        NEWS &         1.09 (0.0) &             \first{1.08} &    0.36 (0.05) &         \first{0.35} \\
        \hline 
    \end{tabular}}
    \caption{Comparison of the HSIC metric on the factual observational dataset and the Augmented dataset obtained by \our. We copy over the CF error from our main table for reference.  We present the $p$ values within brackets for the factual case for the one-sided paired T test with \our\ as the baseline.}\label{tab:hsic_obs}
\end{table*}} 

\end{document}